\documentclass[runningheads]{llncs}
\usepackage{graphicx}

\usepackage{tikz}
\usepackage{comment}
\usepackage{amsmath,amssymb} 
\usepackage{bbding}
\usepackage{color}
\usepackage{booktabs}
\usepackage{bbm}
\usepackage{diagbox}
\usepackage{wrapfig}
\usepackage{pifont}
\usepackage{multirow}
\usepackage[abs]{overpic}
\usepackage{algorithm}
\usepackage[noend]{algpseudocode}
\usepackage{algorithmicx}

\newcommand{\eg}{\textit{e}.\textit{g}.}
\newcommand{\etal}{\textit{et al}.}

\usepackage[accsupp]{axessibility}  
\usepackage{lipsum}

\begin{document}
\pagestyle{headings}
\mainmatter

\title{W2N: Switching From Weak Supervision to Noisy Supervision for Object Detection} 

\titlerunning{W2N}
%
\author{Zitong Huang\inst{1} \and
Yiping Bao\inst{2} \and
Bowen Dong\inst{1} \and Erjin Zhou\inst{2} \and Wangmeng Zuo\inst{1,3}\textsuperscript{\Envelope} } 
\authorrunning{Huang \etal}
%

\institute{
$^1$Harbin Institute of Technology \quad $^2$MEGVII Technology \\ \quad $^3$Peng Cheng Laboratory \\
\email{\{zitonghuang99,cndongsky\}@gmail.com,\{baoyiping,zhouerjin\}@gmail.com, wmzuo@hit.edu.cn}}

\maketitle

\newcommand\blfootnote[1]{%
\begingroup
\renewcommand\thefootnote{}\footnote{#1}%
\addtocounter{footnote}{-1}%
\endgroup
}
\begin{figure*}[t]

    \begin{center}
    \includegraphics[width=1.0\linewidth]{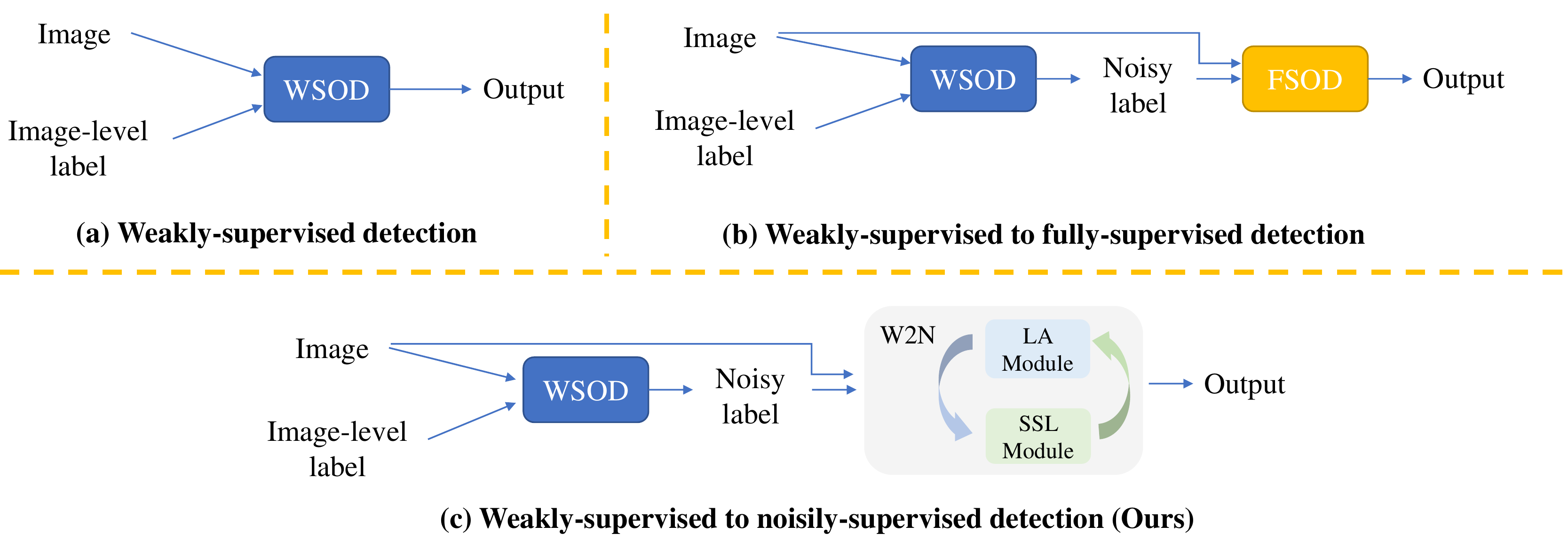}
    \end{center}

       \caption{Training paradigms with three different weakly supervised object detection frameworks: (a) Basic weakly-supervised detection. (b) Weakly-supervised to fully-supervised detection framework. (c) Our W2N framework.}

    \label{fig:smallpipeline}
    \end{figure*}

\begin{abstract}
Weakly-supervised object detection (WSOD) aims to train an object detector only requiring the image-level annotations. 
Recently, some works have managed to select the accurate boxes generated from a well-trained WSOD network to supervise a semi-supervised detection framework for better performance. 
However, these approaches simply divide the training set into labeled and unlabeled sets according to the image-level criteria, 
such that sufficient mislabeled or wrongly localized box predictions are chosen as pseudo ground-truths, resulting in a sub-optimal solution of detection performance. 
To overcome this issue, we propose a novel WSOD framework with a new paradigm that switches from weak supervision to noisy supervision (W2N). 
Generally, with given pseudo ground-truths generated from the well-trained WSOD network, we propose a two-module iterative training algorithm to refine pseudo labels and supervise better object detector progressively. 
In the localization adaptation module, we propose a regularization loss to reduce the proportion of discriminative parts in original pseudo ground-truths, obtaining better pseudo ground-truths for further training. 
In the semi-supervised module, we propose a two tasks instance-level split method to select high-quality labels for training a semi-supervised detector. 
Experimental results on different benchmarks verify the effectiveness of W2N, and our W2N outperforms all existing pure WSOD methods and transfer learning methods. Our code is publicly available at \url{https://github.com/1170300714/w2n\_wsod}.

\keywords{weakly supervised learning, object detection}
\end{abstract}
\section{Introduction}
Different from fully supervised object detection (FSOD)~\cite{girshickICCV15fastrcnn,renNIPS15fasterrcnn} which heavily relys on instance-level bounding box annotations, weakly supervised object detection (WSOD) aims to use only image-level labels as supervision to train an object detector. Compared to the time-consuming instance-level ground-truth annotating process, image-level category labels are easy to obtain relatively, which is more time-saving and economy. Therefore, WSOD has  become a hot and meaningful research topic. 
Existing WSOD methods~\cite{bilen2016weakly,tang2017multiple,tang2018pcl,ren-wetectron2020,cinbis2016weakly} usually follow the multiple instance learning (MIL) framework, which is based on precomputed region proposals~\cite{uijlings2013selective} and is formulated as a proposals classification task, as shown in Fig.~\ref{fig:smallpipeline}~(a). However, without accurate bounding box ground-truths, the localization ability of model is severely limited by inaccurate region proposals. Specifically, the WSOD network tends to focus on the discriminative part instead of the whole object for some typical categories (person, cat, dog, etc.). As shown in Fig.~\ref{fig:smallpipeline}~(b), some works \cite{zhang2018w2f,li2016weakly,tang2017multiple,tang2018pcl,Wan_2018,yang2019towards} proposed pseudo ground-truth (PGT) excavation algorithm to generate pseudo ground-truths from prediction by a MIL-based weakly-supervised object detector and use it to deploy a supervised detector, trying to apply the FSOD training paradigm to WSOD task. However, the improvement of detection precision is still limited because some low-quality boxes in the pseudo ground-truths make the WSOD network converge to the sub-optimal solution. 

To reduce the negative effect from low-quality pseudo ground-truths, some semi-supervised learning~\cite{liu2020unbiased,sohn2020simple} approaches  have been proposed and applied into weakly supervised object detection tasks. \eg, the recently proposed SoS~\cite{sui2021salvage} combines a novel labeled-unlabeled dataset split method as well as the state-of-the-art semi-supervised detection method~\cite{liu2020unbiased} into the WSOD training to improve the detection performance. The main idea of this method is paying more attention to relatively high-quality pseudo labels and carry out a dynamic label updating for noisy labels to improve the performance of detector progressively. 

Inspired by this semi-supervised learning formula, we argue that the pseudo ground-truths can been seen as an inaccurate instance-level bounding box annotation, 
so it's significant to formulate the multi-phase WSOD problem as a noisy-label object detection task. 
To this end, we propose our novel weakly supervised object detection framework namely Weakly-supervision to Noisy-supervision (W2N).  The noisy labels of the training image set are generated by any well-trained WSOD and then fed into W2N framework for further training procedure. An overview of the contrast between the existing WSOD framework and our framework is presented in Fig.~\ref{fig:smallpipeline}~(c).

We formulate W2N framework to an iterative refinement process including several \textbf{localization adaptation modules} and \textbf{semi-supervised learning modules}. In the localization adaptation module, we initialize a fully supervised detector training on the noisy dataset generated by WSOD. During the training phase, we generate a proposal outside each noisy box annotation and then store the decoded boxes of their regression results. Meanwhile, the decoded boxes are used to calculate a regularization loss to optimize the detector. After training, we use this detector to generate pseudo ground-truth again to reduce the  proportion of bounding  box located at discriminative part and then step into the semi-supervised learning module. And in the semi-supervised learning module, we first split the dataset with pseudo ground-truths into labeled set and a unlabeled set by the hybrid-level dataset split method. And then a semi-supervised object detection framework is performed to train a detector on these two sets. Finally, we execute these two modules iteratively and construct an iterative training framework for better detection performance with only image-level annotations. 

Extensive experiments and ablation studies have been conducted to evaluate the effectiveness our proposed method. 
The experimental results demonstrate that our W2N framework brings huge improvement for all baselines on different benchmark datasets. In conclusion, the contributions of this paper are summarized as follows:
\begin{itemize}

    \item [1)] We propose a new multi-phase WSOD paradigm, which formulates the multi-phase weakly supervised object detection problem as a noisy-label object detection problem to reduce the negative effect from low-quality pseudo ground-truths.
   \item [2)] To tackle the noisy-label training problem, we proposed an iterative learning framework including localization adaptation module and semi-supervised learning module, which improves the quality of pseudo ground-truths and the performance of detector.
   \item [3)] Experimental results on different benchmark datasets show that our proposed method bring a huge improvement for all WSOD baseline and achieve state-of-the-art performance on WSOD tasks. 
\end{itemize}

\begin{figure*}[t]

    \begin{center}
    \includegraphics[width=1.0\linewidth]{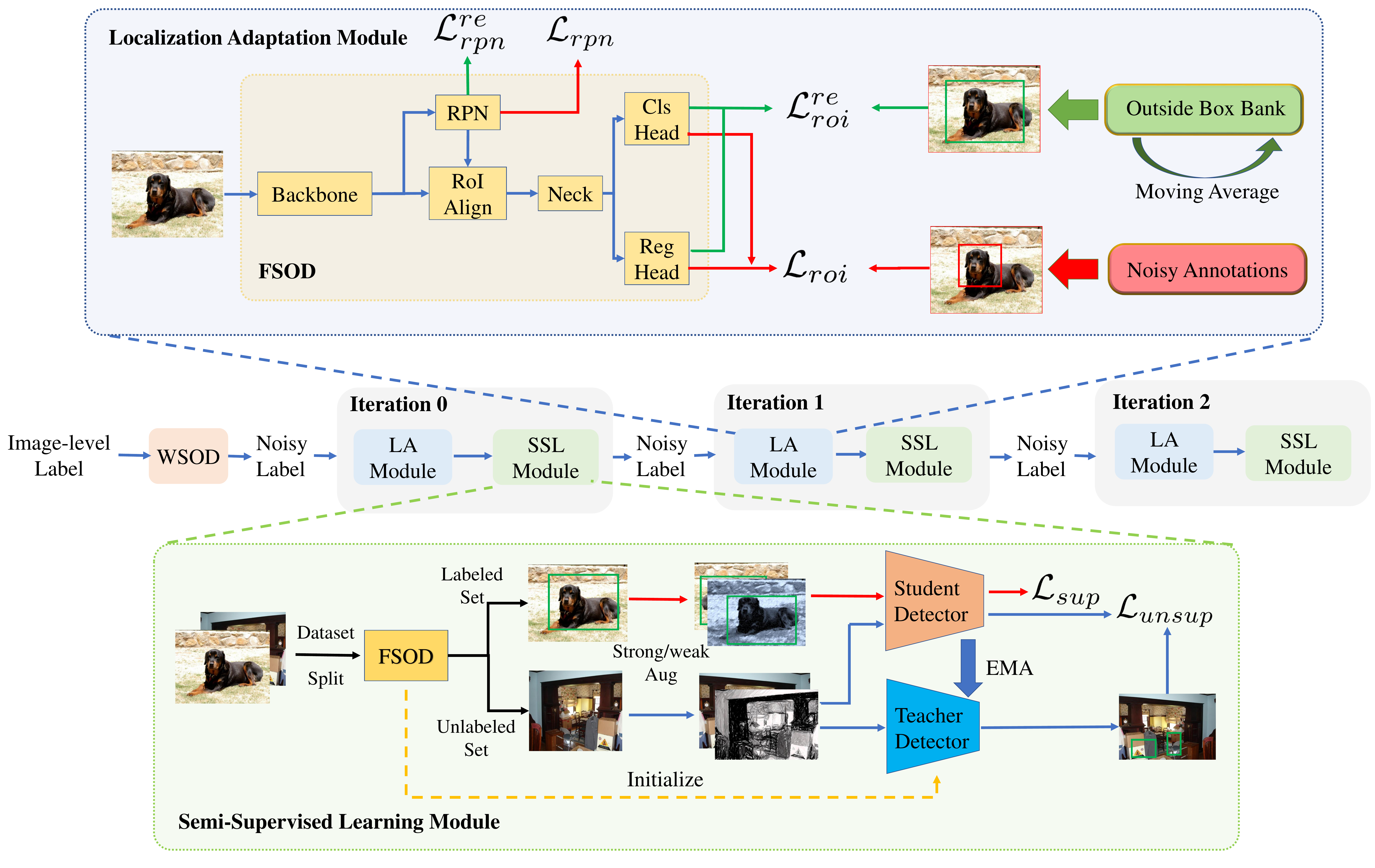}
    \end{center}

       \caption{The illustration of our Weak-to-Noisy (W2N) method, which executes localization adaptation modules (LA module) and semi-supervised learning modules (SSL module) iteratively to generate more accurate  pseudo  labels  and  supervise  a  better  object  detector. Specifically, the localization adaptation module focus on handling bounding boxes of discriminative parts in $\mathbb{X}_{p}$ to enlarge the corresponding bounding box and cover more parts of the object, and the semi-supervised learning module  leverages the pseudo ground-truth of $\mathbb{X}_{p}$ with higher detection precision to enhance the final detection performance.}

    \label{fig:pipeline}
    \end{figure*}
\section{Related Work}

\subsection{Weakly Supervised Object Detection}
Existing WSOD methods~\cite{bilen2016weakly,tang2017multiple,tang2018pcl,Shen_2019_CVPR} are usually based on multiple instance learning (MIL)~\cite{Dietterich1997SolvingTM}, which formulate this task as a proposal classification problem.
%
%
%
Nevertheless, most of the WSOD algorithms tend to recognize the discriminative parts of some objects and optimizing into local-minima, which promote the proposals of several approaches~\cite{Chen2020SLVSL,huangCASD2020,ren-wetectron2020}.
%
%
%
Recently, some works~\cite{dong2021boosting,zhong2020boosting,Cao_2021_ICCV} have leveraged transfer learning paradigm with an external fully-annotated source dataset to further improve the detection performance of WSOD. 
%
%
In addition, some work managed to convert weak supervision into other paradigms. For example, W2F~\cite{zhang2018w2f} combined the weakly-supervised detector and the fully-supervised detector by our pseudo ground-truth mining algorithm. SoS~\cite{sui2021salvage} harness all potential supervisory signals in WSOD and split the dataset into labeled and unlabeled images to execute a SSOD framework. To the best of our knowledge, we are the first to formulate the weakly supervised object detection problem as a noisy-label object
detection problem. In addition, we explore the noise characteristic of every instance-level annotation and design two learning modules to enhance their accuracy, which is not explored in previous works.

\subsection{Learning with Noisy Labels}
Some work are engaged in exploring how to train an image classifier with noisy labels. To address this problem, DivideMix \cite{li2019dividemix} used two networks to perform sample selection via a two-component mixture model. Pleiss \etal \cite{NEURIPS2020_c6102b37} introduced the \textit{Area Under the Margin} statistic which measures the average difference between the logit of a sample’s assigned class and its highest non-assigned to separate correctly-labeled data from mislabeled data. Liu \etal \cite{liu2020early} found that model learns to predict the true labels during the early learning stage but eventually memorizes the wrong labels, which inspires them to leverage the early output of the model. We absorb the inspiration of these work and adapt them to noisy label object detection framework. 

\subsection{Semi Supervised Object Detection}
Semi-supervised learning aims to training networks with both a few of labeled and amount of unlabeled data. In this setting, Jeong \etal \cite{jeong2019consistency} proposed a consistency-based method, which enforces the predictions of an input image and its flipped version to be consistent. STAC \cite{sohn2020simple} proposes to use a weak data augmentation for model training and a strong data augmentation is used for performing pseudo-label. Liu \etal \cite{liu2020unbiased} proposed a simple yet effective method, Unbiased Teacher, to address the pseudolabeling bias issue caused by class-imbalance existing in ground-truth labels and the overfitting issue caused by the scarcity of labeled data.  Xu.\etal \cite{xu2021end} proposed a soft teacher mechanism as well as a box jittering approach to improve the overall detection performance with semi-supervised manner. 

\section{Proposed Method}
\noindent\textbf{Definition. } Let $\mathbb{X} = \{(\mathbf{I},\mathbb{P}, \mathbf{y})\}$ denotes the weakly annotated dataset including $C$ individual object categories, where $\mathbf{I}$ means the input image, $\mathbb{P}$ means the set of proposals w.r.t. $\mathbf{I}$, and $\mathbf{y}= [y_1,y_2,\ldots,y_C]^T$ is the image classification label. WSOD targets at learning an object detector $g$ with only image-level supervision. 
\subsection{Overview}
With given dataset $\mathbb{X}$, we first train a weakly supervised object detector $g$ following previous state-of-the-art methods~\cite{ren-wetectron2020,huangCASD2020,dong2021boosting,tang2017multiple} and then adopt the multi-phase training strategy~\cite{zhang2018w2f} to generate pseudo ground-truth (PGT) on the training images. Now we obtain a new dataset with supervised signal: $\mathbb{X}_{p} = \{(\mathbf{I},\{\mathbf{S}\})\}$ , $\mathbf{S} = (\mathbf{b},c)$, where $\mathbf{b} = [x,y,w,h]$ denotes the instance-level bounding box by its center coordinate $(x,y)$, width $w$, height $h$, and $c$ denotes the category of this box. We propose $\mathbb{X}_{p}$ can be regarded as a noisy annotation due to the low accuracy in terms of classification or localization,  and the WSOD task can be converted to an object detection task with noisy annotations. To train an object detector on such noisy dataset, we propose a novel training framework namely Weakly-to-Noisy (W2N),  which executes localization adaptation modules and semi-supervised learning modules iteratively to generate more accurate pseudo labels and supervise a better object detector. The overall pipeline of W2N is shown as Fig.\ref{fig:pipeline}. Specifically, the localization adaptation module focus on handling discriminative parts bounding box in $\mathbb{X}_{p}$ to enlarge the corresponding bounding box and cover more parts of the object, and the semi-supervised learning module  leverages the high-quality part of the pseudo ground-truths in $\mathbb{X}_{p}$ to enhance the final detection performance of object detector.
\subsection{Noisy Label Generation}
Due to the lack of instance-level supervision during the training procedure of WSOD, the prediction results from the pretrained WSOD network $g$ is not accurate enough~\cite{bilen2016weakly,tang2017multiple,zhang2018w2f}, \eg, the wrong prediction in Fig.~\ref{fig:in}, mislabel or low location accuracy.
Following \cite{zhang2018w2f,sui2021salvage}, we treat the pretrained object detector $g$ as a generator of noisy labels to generate the pseudo ground-truths.
We select three WSOD baseline methods to play the role of generators: \emph{OICR+REG} \cite{tang2017multiple}, \emph{CASD}\cite{huangCASD2020}, and \emph{LBBA}\cite{dong2021boosting}.
After training on $\mathbb{X}$, the weakly-supervised detector $g$ inference on training Image $\mathbf{I}$  and we filter the original predictions, convert it to pseudo ground-truth and obtain $\mathbb{X}_{p}$ according to the Pseudo Ground-Truth Excavation method proposed by  W2F\cite{zhang2018w2f}.

\subsection{Learning Detector with Noisy Annotations}

After generating the noisy labels, we feed the labels into the W2N training framework to supervise better object detector progressively. Following \cite{li2019dividemix,liu2020early}, we propose an training framework W2N, which iterates between {localization adaptation module} and {semi-supervised learning module} for several steps. The following subsections will illustrate these two modules in details.
\begin{figure*}[t!]
    \begin{center}
    \includegraphics[width=4.7in]{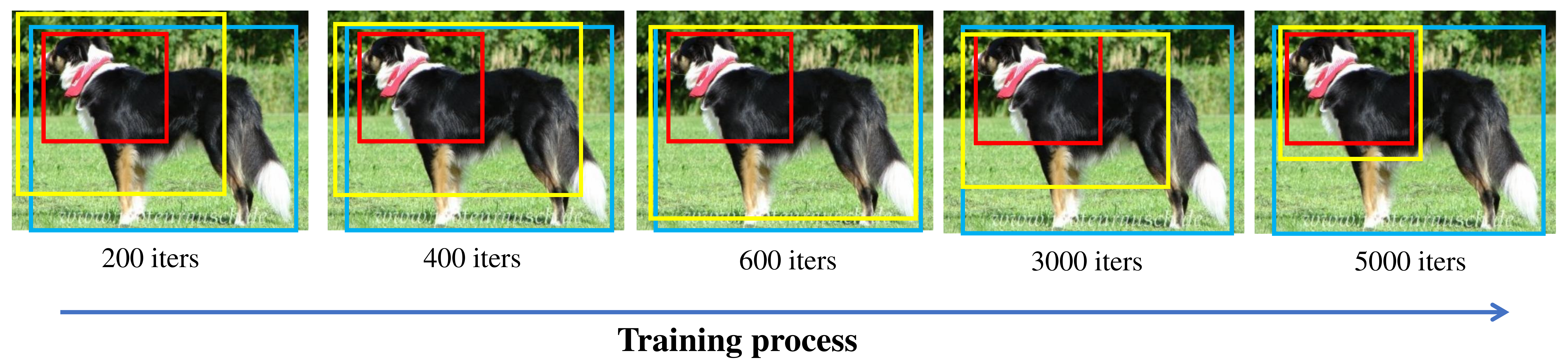}
    \end{center}

       \caption{An example of regression results of a proposals outside the discriminative part pseudo ground-truth during training. Blue box indicates the real ground-truth, red box indicates the discriminative part pseudo ground-truth and the yellow box indicates the outer box of the red one.  Yellow box is regressed to the blue box at early stage of training process, but finally overfits to the red box.}

    \label{fig:el}
    \end{figure*}
    
\subsubsection{Localization Adaptation Module.}
\label{sec:lam}
\begin{figure*}[t!]

    \centering
    \includegraphics[scale=0.35]{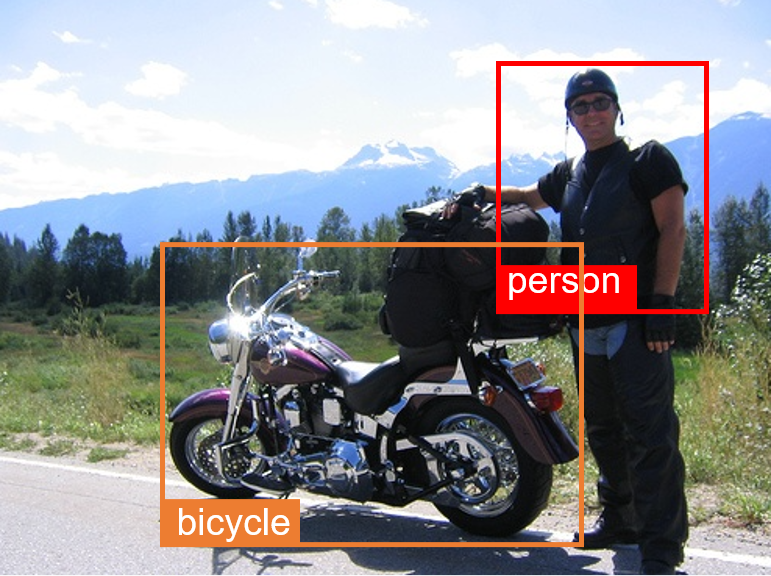}
    \centering

       \caption{An example of noisy label. Notice that the orange box has precise bounding box but mislabeled to bicycle (the ground-truth is motorbike), while the category of red box is correct but its bounding box is incorrect.}

    \label{fig:in}
    \end{figure*}

In semi-supervised learning module which will be mentioned below, the quality of labeled set will effect the performance of the detector \cite{sui2021salvage}. The more accurate label in labeled set, the higher performance the model achieve. However, we argue that the dataset split can not recognize and filter the discriminative-part noisy labels among several categories (e.g., like the ``person'' prediction box Fig~\ref{fig:in}.  
The main reason is that too many discriminative-part noisy labels appear in the $\mathbb{X}_{p}$ such that network tends to overfit them easily during training and then obtain low detection precision.

To deal with this problem, we revisit the characteristic of discriminative-part noisy labels and dig out such regular pattern, which is shown in Fig. \ref{fig:el}. First, the discriminative-part noisy labels are usually inside the corresponding real ground-truths. Second, if we use the $\mathbb{X}_{p}$ to train a supervised object detector $f$, the outer proposals of the discriminative part noisy labels will regress toward the real ground-truth during the early stage during of training phase. But as training continues, it tends to overfit toward the discriminative part noisy labels again. Based on this observation, we refer to the method of using early output in noisy-label image classification task and design a regularization loss to handle the ``discriminative part problem''.

As mentioned above, with regard to a discriminative part noisy labels, their corresponding outer proposals will regress toward a more accurate location at early stage learning phase. Therefore we store these proposals as the extra supervision to optimize the fully supervised detector $f$. Specifically, given a pseudo ground-truth box $\mathbf{b} = [x,y,w,h]$ at iteration $t$ during training phase, we randomly generate a outer box extending from $\mathbf{b}$ it by random sampling the transformation $\delta^{t}$:
\begin{equation}
    \begin{split}
        \delta^{t}_{x}, \delta^{t}_{y} &\sim \mathcal{U}(-\alpha,\alpha) \\
        \delta^{t}_{w}, \delta^{t}_{h} &\sim \mathcal{U}(\sqrt{3},2) 
    \end{split}
\end{equation}
$\mathcal{U}(-\alpha,\alpha)$ denotes an uniform distribution in the range $[-\alpha, \alpha]$. Then a random outer box $\tilde{\mathbf{b^{t}}}=[\tilde{x^{t}},\tilde{y^{t}},\tilde{w^{t}},\tilde{h^{t}}]$ is obtained by:
\begin{equation}
    [\tilde{x^{t}}, \tilde{y^{t}}, \tilde{w^{t}}, \tilde{h^{t}}] = [x+\delta^{t}_x \cdot w, y+\delta^{t}_y \cdot h, w \cdot \delta^{t}_w, h \cdot \delta^{t}_h].
\end{equation}
The outer boxes $\tilde{\mathbf{b^{t}}}$ are fed into the object detector and then obtain the decode boxes $\hat{\mathbf{b^{t}}}$. To measure the quality of $\hat{\mathbf{b^{t}}}$, we only select the boxes whose prediction scores are higher than a threshold $\tau_{score}$ while the  IoU with corresponding $\mathbf{b}$ are lower than the label assigning threshold $\tau_{assign}$ (e.g., 0.5).  Finally, to obtain more precision outer boxes, we adopt the moving average strategy to synthesize all $\hat{\mathbf{b}}$ before iteration $t$ and obtain the extra supervision for regularization, shown as Eqn.~(\ref{boxema}):
\begin{equation}\label{boxema}
    \hat{\mathbf{b}}^{t}_{re} = \beta\hat{\mathbf{b}}^{t-1}_{re}+(1-\beta)\hat{\mathbf{b^{t}}},
\end{equation}
where $\beta$ is the moving average value of bounding box. Then we use  $\{(\mathbf{b},c)\}$ and $\{(\hat{\mathbf{b}}^{t}_{re},c)\}$ as the supervision signal to optimize detector $f$, and calculate loss function $\mathcal{L}_{rpn}$, $\mathcal{L}_{roi}$, $\mathcal{L}^{re}_{rpn}$ and $\mathcal{L}^{re}_{roi}$, where $\mathcal{L}_{rpn}$ and $\mathcal{L}_{roi}$ indicate the loss supervised with noisy labels $\{(\mathbf{b},c)\}$ of RPN and RoI head while  $\mathcal{L}^{re}_{rpn}$ and $\mathcal{L}^{re}_{roi}$ is calculated with extra supervision $\{(\hat{\mathbf{b}}^{t}_{re},c)\}$ as regularization terms. Each of them is the combination of Smooth L1 Loss(regression loss) and Cross-Entropy Loss(classification loss), which are the same formulation as \cite{renNIPS15fasterrcnn}. The whole loss function  $\mathcal{L}_{fsod}$ for optimization $f$ is shown as Eqn.~(\ref{equ:fsod}):
\begin{equation}\label{equ:fsod}
    \mathcal{L}_{fsod} = \mathcal{L}_{rpn}+\mathcal{L}_{roi}+\lambda_{re}(\mathcal{L}^{re}_{rpn}+\mathcal{L}^{re}_{roi})
\end{equation}
where $\lambda_{re}$ indicates the regularization weight.

After the process above, we use the well-trained detector $f$ to re-generate the pseudo ground-truths on the training set, which can reduce the proportion of low-quality pseudo ground-truths and improve the performance of the next semi-supervised learning module. 

\subsubsection{Semi-Supervised Learning Module.}
In this module, we design a hybrid-level dataset split algorithm as well as a pseudo-label based semi-supervised training algorithm. 

Dataset split method is crucial for turning noisy-label learning into semi-supervised approach.  A basic solution is that spliting the whole dataset according to the training loss of each image.
The training data with small loss is regarded as the sample from labeled set, vise versa. SoS \cite{sui2021salvage} proposed the ``image-level split method'', which accumulated the losses from the RPN module and that from the detection head and then obtained the image-level split loss function. Given image $\mathbf{I}$, the image-level split loss $\mathcal{L}_{split}(\mathbf{I})$ is defined as Eqn.~(\ref{eqn:image-split}): 
\begin{equation}\label{eqn:image-split}
\begin{split}
      \hspace{-4mm}\mathcal{L}_{split}(\mathbf{I})=\mathop{\text{avg}}\limits_{i}(\mathcal{L}^{rpn}_{split}(R_i,t_i))+\mathop{\text{avg}}\limits_{j}(\mathcal{L}^{roi}_{split}(R_j,t_j)).
\end{split}
\end{equation}
And the $\mathcal{L}^{rpn}_{split}$ and $\mathcal{L}^{roi}_{split}$ are shown as Eqn.~(\ref{eq6}) and (\ref{eq7}):
\begin{equation}\label{eq6}
    \mathcal{L}^{rpn}_{split}(R_i,t_i)=\mathcal{L}^{cls}_{rpn}(R_i,t_i)+\mathcal{L}^{reg}_{rpn}(R_i, t_i),
\end{equation}

\begin{equation}\label{eq7}
    \mathcal{L}^{roi}_{split}(R_j,t_j)=\mathcal{L}^{cls}_{roi}(R_j,t_j)+\mathcal{L}^{reg}_{roi}(R_j,t_j), 
\end{equation}
where $R_i$ is the i-th foreground RoI, $t_i$ indicates the assigned target label of $R_i$, $\mathcal{L}_{rpn}$ and $\mathcal{L}_{roi}$ are RPN and RoI head losses, and $cls$ and $reg$ stand for classification task and box regression task, respectively. $\mathcal{L}^{cls}_{*}$ is Cross-Entropy Loss and $\mathcal{L}^{reg}_{*}$ is Smooth L1 Loss. And the \text{avg}($\cdot$) means the mean average operation.
Then, we rank all instances with their $\mathcal{L}_{split}(\textbf{I})$ by the ascending order, keeping the number of $p$ percent of  image annotations with small loss value to be the labeled set. However, we find that a training image may contain multiple instance labels, and the accurate labels and noisy labels often appear at the same time.  Therefore, we proposed the second split method namely ``instance-level split method'', in which every instance is be seen to the smallest split unit. And the aggregated loss in Eqn.~(\ref{eqn:image-split}) will be modified to Eqn.~(\ref{eqn:instance-split}):
\begin{equation}\label{eqn:instance-split}
\begin{split}
      \hspace{-4mm}\mathcal{L}_{split}(\textbf{S})=\mathop{\text{avg}}\limits_{i}(\mathcal{L}^{rpn}_{split}(R_i,t_i))+\mathop{\text{avg}}\limits_{j}(\mathcal{L}^{roi}_{split}(R_j,t_j)),    
\end{split}
\end{equation}
where $\textbf{S}$ indicates to one instance label 
and \text{avg}($\cdot$) means the mean average operation. 
Then we rank all instances according to the $\mathcal{L}_{split}(\textbf{S})$ by the ascending order, and then keep the top $p$ percent of the instance labels with small loss value to be the labeled set $\mathbb{X}_{l} = \{(\mathbf{I}_{l},\{\mathbf{S}_{l}\})\}$, and
the other instances are keeping unlabeled. 

In SoS \cite{sui2021salvage}, the labeled set are used for supervising the training for classification and regression sub-tasks. 
However, we can not make sure that each pseudo label is correct in terms of both classification and localization. As shown in Fig.~\ref{fig:in}, a box with high location information may be mislabeled of category while a box with correct category may cover part of an object. From this perspective, we introduce two tags $\lambda_{cls},\lambda_{reg}$ for one instance label indicating their confidence for two sub-task respectively. The final formulation of labeled set is modified to $\mathbb{X}_{l} = \{(\mathbf{I}_{l},\{(\mathbf{S}_{l}, \lambda_{cls},\lambda_{reg})\})\}$, where $\lambda_{cls}, \lambda_{reg} \in \{0,1\}, \lambda_{cls}+\lambda_{reg} \neq 0$. $\lambda_{cls}=1$ means the category label of this instance is correct, while $\lambda_{cls}=0$ means not, similar meaning  for $\lambda_{reg}$. To decide the value of  $\lambda_{cls},\lambda_{reg}$, we propose ``two tasks instance-level split'' method, which is shown as Eqn.~(\ref{eq:twotasksplit}):
\begin{equation}\label{eq:twotasksplit}
\begin{split}
    \hspace{-4mm}\mathcal{L}^{cls}_{split}(\textbf{S}) = \mathop{\text{avg}}\limits_{i}(\mathcal{L}^{cls}_{rpn}(R_i,t_i))+\mathop{\text{avg}}\limits_{j}(\mathcal{L}^{cls}_{roi}(R_j,t_j)),\\
    \hspace{-4mm}\mathcal{L}^{reg}_{split}(\textbf{S}) = \mathop{\text{avg}}\limits_{i}(\mathcal{L}^{reg}_{rpn}(R_i,t_i))+\mathop{\text{avg}}\limits_{j}(\mathcal{L}^{reg}_{roi}(R_j,t_j)),
\end{split}
\end{equation}
where $\mathcal{L}^{cls}_{split}(\textbf{S})$ only accumulates the classification loss for each foreground proposal while $\mathcal{L}^{reg}_{split}(\textbf{S})$ only accumulates the regressions loss for each foreground proposal. Then, we rank the instance according to $\mathcal{L}^{cls}_{split}(\textbf{S})$ and $\mathcal{L}^{reg}_{split}(\textbf{S})$ by the ascending order, respectively. 
Finally, we set $\lambda_{cls}=1$ for the top $p$ percent of the instances in terms of $\mathcal{L}^{cls}_{split}(\textbf{S})$ and set $\lambda_{reg}=1$ for the top $p$ percent of the instances in terms of $\mathcal{L}^{reg}_{split}(\textbf{S})$.
In Sec.~{\ref{sec:exp}}, we will discuss the effect of three data split proposed above.

After spliting the noisy dataset, we introduce a novel semi-supervised object detection method for weakly-to-noisy label training.
The difference between \cite{liu2020unbiased} and our semi-supervised detection method is two-fold. 
First, we use labeled set $\mathbb{X}_{l}$ as labeled set to optimize model with the supervised loss $\mathcal{L}_{sup}$. Combining with our \emph{two tasks instance-level split} method, we modify the origin supervised loss function with adding the value of $(\lambda_{cls},\lambda_{reg})$. Specifically, $\mathcal{L}_{sup}$ is shown as Eqn.~(\ref{eqn:ssod-sup}):
\begin{equation}\label{eqn:ssod-sup}
\begin{split}
    \hspace{-2mm}\mathcal{L}_{sup}(\text{I}) &= \mathop{\text{avg}}\limits_{i}(\lambda^{t_i}_{cls}\mathcal{L}^{cls}_{rpn}(R_i, t_i)+\lambda^{t_i}_{reg}\mathcal{L}^{reg}_{rpn}(R_i, t_i))\\
    &+\mathop{\text{avg}}\limits_{j}(\lambda^{t_j}_{cls}\mathcal{L}^{cls}_{roi}(R_j, t_j)+\lambda^{t_j}_{reg}\mathcal{L}^{reg}_{roi}(R_j, t_j))\\
    &+ \mathop{\text{avg}}\limits_{k} (\mathcal{L}_{bg}(R_k)),
\end{split}
\end{equation}
where $\mathcal{L}_{bg}(R_i)$ indicates the background loss of corresponding proposals. Particularly, only the target label of which $\lambda^{t_i}_{cls}=1$ ($\lambda^{t_i}_{reg}=1$) can contribute to $\mathcal{L}_{sup}$ in classification (regression) task. The loss function used on the labeled set is shown as Eqn.~(\ref{eqn:ssod-sup-all}):
\begin{equation}\label{eqn:ssod-sup-all}
    \mathcal{L}_{sup}=\frac{1}{N_{l}}\sum_{i}\mathcal{L}_{sup}(\text{I}_{i}),
\end{equation}
where $N_{l}$ is the number of image in $\mathbb{X}_{l}$. Second, the regression loss of the unlabeled data are not adopted in the whole training process of \cite{liu2020unbiased}. In our method we adopt the box jittering strategy proposed by \cite{xu2021end} and add the regression loss of the unlabeled data in origin $\mathcal{L}_{unsup}$ \cite{liu2020unbiased}. Finally, the whole loss function of SSOD module is shown as Eqn.~(\ref{eq:losstotal}):
\begin{equation}\label{eq:losstotal}
    \mathcal{L}_{ssod}=\mathcal{L}_{sup}+\lambda_{u}\mathcal{L}_{unsup},
\end{equation}
where $\lambda_{u}$ is the weight of $\mathcal{L}_{unsup}$.
\subsubsection{Iterative Training Framework}
Finally, we propose the two-phase iterative training framework based on these two modules. The whole training process of our framework is given in Algorithm \ref{alg:w2n}, which is summarized as follows. Specifically, the first phase is the conventional weakly-supervised object detection pretraining module, we train a WSOD network $g$ and then generated the pseudo ground-truths for each training image in the training dataset $\mathbb{X}_{p}^{0}$. The second phase is our proposed weakly-to-noisy training framework. Given the pseudo ground-truths, we first execute the localization adaptation module to initialize a fully-supervised detector $f_t$ and then refine $\mathbb{X}_{p}^{t}$ to reduce the proportion of the discriminative part. Then we excute the two tasks instance-level split method and split the whole training set $\mathbb{X}_{p}^{t}$ into labeled set and unlabeled set. With the splitted training sets, we execute the semi-supervised object detection module to supervise a better object detector $f^{'}_{t}$. Generally, we use $f^{'}_{t}$ to update the $\mathbb{X}_{p}^{t}$ to $\mathbb{X}_{p}^{t+1}$ and then perform these two modules iteratively for $T$ times. And finally, the last object detector $f^{'}_{T}$ with corresponding parameters $\theta_{f}^{T}$ is saved for usage. 

\begin{algorithm}[t]

   \caption{Weak Supervision to Noisy Supervision for Object Detection}
   \label{alg:w2n}
   \begin{algorithmic}[1]
      \Require Iteration number $T$, weakly annotated dataset  $\mathbb{X}$;
      \Ensure An updated detector $f^{'}_T$;
      \State Train the weakly supervised detector $g$ on $\mathbb{X}$;
      \State Obtain the noisy annotations dataset $\mathbb{X}^{0}_{p}$ by pretrained weakly supervised detector $g$;
      \For{${t}=0...{T-1}$}
      \State \textbf{Localization Adaptation module:}
         \State Initialize an object detector $f_{t}$ on $\mathbb{X}^{t}_{p}$;
         \State Refine $\mathbb{X}^{t}_{p}$ by $f_{t}$;
      \State \textbf{Semi-Supervised Learning module:}
         \State Split $\mathbb{X}^{t}_{p}$ into labeled set and unlabeled set by $f_{t}$;
         \State Execute the semi-supervised object detection approach to optimize $f_t$ to $f^{'}_t$ ; 
         \State Update the $\mathbb{X}^{t}_{p}$ to $\mathbb{X}^{t+1}_{p}$ by $f^{'}_t$ ;
      \EndFor
      
   \end{algorithmic}
   
\end{algorithm}

\begin{table*}[t]
    \caption{Comparison of our method on PASCAL VOC 2007 test set to  state-of-the-art WSOD methods in terms of mAP ($\%$), where $\text{~}^{+}$ means the results with multi-scale testing.}
    \centering
    \resizebox{\textwidth}{!}{
    {
    \begin{tabular}{l | c c c c c c c c c c c c c c c c c c c c | c}
    \specialrule{.15em}{.05em}{.05em}
    Methods & Aero & Bike & Bird & Boat & Bottle & Bus & Car & Cat & Chair & Cow & Table & Dog & Horse & Motor & Person & Plant & Sheep & Sofa & Train & ~~TV~~ & ~~AP~~ \\
    \hline
    \textbf{Pure WSOD}:\\
    WSDDN~\cite{bilen2016weakly} &39.4 & 50.1 & 31.5 & 16.3 & 12.6 & 64.5 & 42.8 & 42.6 & 10.1 & 35.7 & 24.9 & 38.2 & 34.4 & 55.6 & 9.4 & 14.7 & 30.2 & 40.7 & 54.7 & 46.9 & 34.8 \\
    $\text{OICR}^{+}$ \cite{tang2017multiple} & 58.0 & 62.4 & 31.1 & 19.4 & 13.0 & 65.1 & 62.2 & 28.4 & 24.8 & 44.7 & 30.6 & 25.3 & 37.8 & 65.5 & 15.7 & 24.1 & 41.7 & 46.9 & 64.3 & 62.6 & 41.2 \\
    $\text{PCL}^{+}$ \cite{tang2018pcl} & 54.4 & 69.0 & 39.3 & 19.2 & 15.7 & 62.9 & 64.4 & 30.0 & 25.1 & 52.5 & 44.4 & 19.6 & 39.3 & 67.7 & 17.8 & 22.9 & 46.6 & 57.5 & 58.6 & 63.0 & 43.5 \\
    $\text{Yang }\textit{et al.}^{+}$ \cite{yang2019towards} & 57.6 & 70.8 & 50.7 & 28.3 & 27.2 & 72.5 & 69.1 & 65.0 & 26.9 & 64.5 & 47.4 & 47.7  & 53.5 & 66.9 & 13.7 & 29.3 & 56.0 & 54.9 & 63.4 & 65.2 & 51.5 \\
    $\text{C-MIDN}^{+}$ \cite{Gao_2019_ICCV} & 53.3 & 71.5 & 49.8 & 26.1 & 20.3 & 70.3 & 69.9 & 68.3 & 28.7 & 65.3 & 45.1 & {64.6} & 58.0 & 71.2 & 20.0 & 27.5 & 54.9 & 54.9 & {69.4} & 63.5 & 52.6 \\
    Arun \textit{et al.} \cite{Arun_2019} & 66.7 & 69.5 & 52.8 & 31.4 & 24.7 & {74.5} & 74.1 & 67.3 & 14.6 & 53.0 & 46.1 & 52.9 & 69.9 & 70.8 & 18.5 & 28.4 & 54.6 & {60.7} & 67.1 & 60.4 & 52.9 \\
    $\text{WSOD2}^{+}$ \cite{Zeng_2019_ICCV} & 65.1 & 64.8 & {57.2} & {39.2} & 24.3 & 69.8 & 66.2 & 61.0 & 29.8 & 64.6 & 42.5 & 60.1 & {71.2} & 70.7 & 21.9 & 28.1 & 58.6 & 59.7 & 52.2 & 64.8 & 53.6 \\
    GradingNet-C-MIL \cite{Jia_Wei_Ruan_Zhao_Zhao_2021} &-&-&-&-&-&-&-&-&-&-&-&-&-&-&-&-&-&-&-&-& 54.3 \\
    MIST-Full \cite{ren-wetectron2020} & {68.8} & {77.7} & 57.0 & 27.7 & {28.9} & 69.1 & {74.5} & 67.0 & {32.1} & {73.2} & 48.1 & 45.2 & 54.4 & {73.7} & {35.0} & 29.3 & {64.1} & 53.8 & 65.3 & {65.2} & {54.9}  \\
    $\text{IM-CFB}^{+}$ \cite{Yin_Deng_Zhou_Li_2021} & 63.3 & 77.5 & 48.3 & 36.0 & 32.6 & 70.8 & 71.9 & 73.1 & 29.1 & 68.7 & 47.1 & 69.4 & 56.6 & 70.9 & 22.8 & 24.8 & 56.0 & 59.8 & 73.2 & 64.6 & 55.8 \\
    CASD \cite{huangCASD2020} &-&-&-&-&-&-&-&-&-&-&-&-&-&-&-&-&-&-&-&-&  56.8 \\
    SoS \cite{sui2021salvage} & 72.9 & 79.4& 59.6 &20.4 &49.8 &81.2 &82.9 &84.0 &31.5& 76.6& 57.4 &60.7& 74.7 &75.1& 33.0& 34.3 &66.3 &61.1& 80.6& 71.8 & 62.7\\
    $\text{SoS}^{+}$ \cite{sui2021salvage} & 77.9 &  81.2  & 58.9  & 26.7 &  54.3 &  82.5  & 84.0 &  83.5 & 36.3 &  76.5  & 57.5 &  58.4  & 78.5  & 78.6  & 33.8  & 37.4  & 64.0 &  63.4 &  81.5 &  74.0 & 64.4 \\

     \textbf{OICR+REG (reproduce)} & 54.0&61.9&43.9&22.6&31.7&73.8&65.1&60.6&14.4&68.0&17.0&48.8&58.3&69.9&12.8&22.0&53.9&53.6&69.7&60.4&48.3\\
    \textbf{CASD (reproduce)} & 68.8 & 67.2 & 53.9& 38.2& 21.5& 70.4& 69.7 & 68.9& 23.6 & 66.3 & 48.8 & 62.3 & 56.4 & 70.6 & 17.3 & 24.9 & 55.9 & 58.9 & 66.0 & 69.1 & 54.0\\
  
    \textbf{OICR+REG+W2N (Ours)} &71.0&74.2&60.8&28.8&44.6&78.0&72.6&80.3&16.7&74.3&24.3&58.2&64.6&75.1&13.3&29.9&60.3&65.3&80.1&67.6&\textbf{57.0(+8.7)}\\
    \textbf{CASD+W2N (Ours)}& 74.0&81.7&71.2&48.9&51.0&78.6&82.3&83.5&29.1&76.9&51.5&82.1&76.9&79.1&28.5&34.3&65.0&64.2&75.2&74.8&\textbf{65.4(+11.4)}\\
    \specialrule{.15em}{.05em}{.05em}
    \textbf{WSOD with transfer learning}:\\
    $\text{MSD-Ens}^{+}$ \cite{msd2018} & 70.5 & 69.2 & 53.3 & 43.7 & 25.4 & 68.9 & 68.7 & 56.9 & 18.4 & 64.2 & 15.3 & 72.0 & 74.4 & 65.2 & 15.4 & 25.1 & 53.6 & 54.4 & 45.6 & 61.4 & 51.1 \\
    OICR+UBBR \cite{ubbr2018} & 59.7 & 44.8 & 54.0 & 36.1 & 29.3 & 72.1 & 67.4 & 70.7 & 23.5 & 63.8 & 31.5 & 61.5 & 63.7 & 61.9 & 37.9 & 15.4 & 55.1 & 57.4 & 69.9 & 63.6 & 52.0 \\
    $\text{LBBA}^{+}$ \cite{dong2021boosting} & 70.3 & 72.3 & 48.7 & 38.7 & 30.4 & 74.3 & 76.6 & 69.1 & 33.4 & 68.2 & 50.5 & 67.0 & 49.0 & 73.6 & 24.5 & 27.4 & 63.1 & 58.9 & 66.0 & 69.2 & 56.6 \\
    $\text{Zhong \textit{et al.} (R50-C4)}^{+}$ \cite{zhong2020boosting} & 64.8 & 50.7 & 65.5 & 45.3 & 46.4 & 75.7 & 74.0 & 80.1 & 31.3 & 77.0 & 26.2 & 79.3 & 74.8 & 66.5 & 57.9 & 11.5 & 68.2 & 59.0 & 74.7 & 65.5 & 59.7 \\
    $\text{TraMaS}^{+}$ \cite{2021arXiv211014191L} & 68.6 & 61.1& 69.6& 48.1 &49.9 &76.3 &77.8& 80.9 &34.9& 77.0& 31.1& 80.9& 78.5& 66.3& 64.0 &19.1 &69.1& 62.3& 74.4& 69.1& 62.9\\
    $\text{CaT}_{5}$ \cite{Cao_2021_ICCV} & 74.0& 70.7& 60.0 &31.1& 50.0& 75.9& 82.0& 70.7& 32.8& 74.3& 69.5& 70.2 &69.5& 77.0& 37.5& 45.8& 67.0 &61.1& 72.4 &68.0 & 63.0\\
     \textbf{LBBA (reproduce)} & 70.2&75.5&49.2&41.9&30.5&80.5&78.2&72.8&36.4&73.8&52.3&67.0&46.4&76.2&34.6&29.4&67.9&66.6&68.3&74.1&59.1\\
    \textbf{LBBA+W2N (Ours)}& 71.8&83.0&69.9&50.3&54.5&79.0&83.9&83.9&39.4&79.2&52.9&82.2&83.6&79.2&62.6&32.7&68.5&66.1&75.8&74.5&\textbf{68.6(+9.5)} \\
    \specialrule{.15em}{.05em}{.05em}
    \textbf{Upper bounds:} \\
    \hline
    Faster R-CNN (Res50+FPN) \cite{renNIPS15fasterrcnn} &82.8&84.2&75.2&62.4&67.0&81.4&87.1&82.6&57.3&82.5&64.9&83.0&84.0&82.7&83.7&54.0&76.1&73.4&81.8&76.1&\textbf{76.1}\\

    \specialrule{.15em}{.05em}{.05em}
    \end{tabular}
    }}
    \label{table:per-cls-voc07}

\end{table*}

\begin{table*}[tbh]
    \caption{Comparison of our method on PASCAL VOC 2007 trainval set to  state-of-the-art WSOD methods in terms of CorLoc ($\%$), where $\text{~}^{+}$ means the results with multi-scale testing.}
    \centering
    \resizebox{\textwidth}{!}{
    {
    \begin{tabular}{l | c c c c c c c c c c c c c c c c c c c c | c}
    \specialrule{.15em}{.05em}{.05em}
    Methods & Aero & Bike & Bird & Boat & Bottle & Bus & Car & Cat & Chair & Cow & Table & Dog & Horse & Motor & Person & Plant & Sheep & Sofa & Train & ~~TV~~ & CorLoc \\
    \hline
    \textbf{Pure WSOD}: \\
    WSDDN~\cite{bilen2016weakly} & 65.1 & 58.8 & 58.5 & 33.1 & 39.8 & 68.3 & 60.2 & 59.6 & 34.8 & 64.5 & 30.5 & 43.0 & 56.8 & 82.4 & 25.5 & 41.6 & 61.5 & 55.9 & 65.9 & 63.7 & 53.5 \\
    $\text{OICR}^{+}$ \cite{tang2017multiple} & 81.7 & 80.4 & 48.7 & 49.5 & 32.8 & 81.7 & 85.4 & 40.1 & 40.6 & 79.5 & 35.7 & 33.7 & 60.5 & 88.8 & 21.8 & 57.9 & 76.3 & 59.9 & 75.3 & {81.4} & 60.6 \\
    $\text{PCL}^{+}$ \cite{tang2018pcl} & 79.6 & 85.5 & 62.2 & 47.9 & 37.0 & 83.8 & 83.4 & 43.0 & 38.3 & 80.1 & 50.6 & 30.9 & 57.8 & 90.8 & 27.0 & 58.2 & 75.3 & {68.5} & 75.7 & 78.9 & 62.7 \\
    $\text{Li}^{+}$ \cite{li2019weakly} & 85.0 & 83.9 & 58.9 & 59.6 & 43.1 & 79.7 & 85.2 & 77.9 & 31.3 & 78.1 & 50.6 & 75.6 & 76.2 & 88.4 & 49.7 & 56.4 & 73.2 & 62.6 & 77.2 & 79.9 & 68.6 \\
    $\text{C-MIL}^{+}$ \cite{Wan_2019} &-&-&-&-&-&-&-&-&-&-&-&-&-&-&-&-&-&-&-&-& 65.0 \\
    $\text{Yang }\textit{et al.}^{+}$ \cite{yang2019towards} & 80.0 & 83.9 & 74.2 & 53.2 & 48.5 & 82.7 & 86.2 & 69.5 & 39.3 & 82.9 & 53.6 & 61.4& 72.4 & 91.2 & 22.4 & 57.5 & {83.5} & 64.8 & 75.7 & 77.1 & 68.0 \\
    $\text{MIST (Full)}^{+}$ \cite{ren-wetectron2020} & 87.5 & 82.4 & {76.0} & 58.0 & 44.7 & 82.2 & 87.5 & 71.2 & {49.1} & 81.5 & 51.7 & 53.3 & 71.4 & 92.8 & 38.2 & 52.8 &79.4 & 61.0 & 78.3 & 76.0  & 68.8 \\
    $\text{WSOD2}^{+}$ \cite{Zeng_2019_ICCV} & 87.1 & 80.0 & 74.8 & 60.1 & 36.6 & 79.2 & 83.8 & 70.6 & 43.5 & {88.4} & 46.0 & {74.7} & 87.4 & 90.8 & 44.2 & 52.4 & 81.4 & 61.8 & 67.7 & 79.9 & 69.5\\
    Arun~ \textit{et al.}\cite{Arun_2019} & {88.6} & {86.3} & 71.8 & 53.4 & 51.2 & {87.6} & {89.0} & 65.3 & 33.2 & 86.6 & 58.8 & 65.9 & {87.7} & {93.3} & 30.9 & 58.9 & 83.4 & 67.8 & 78.7 & 80.2 & {70.9} \\
    GradingNet-C-MIL \cite{Jia_Wei_Ruan_Zhao_Zhao_2021} &-&-&-&-&-&-&-&-&-&-&-&-&-&-&-&-&-&-&-&-&  72.1 \\
    $\text{IM-CFB}^{+}$ \cite{Yin_Deng_Zhou_Li_2021} &-&-&-&-&-&-&-&-&-&-&-&-&-&-&-&-&-&-&-&-& 72.2 \\
     \textbf{OICR+REG (reproduce)} & 91.6&78.3&62.6&46.0&44.8&86.4&87.7&80.3&34.4&87.1&30.1&69.4&81.1&90.8&31.3&44.8&76.0&76.1&83.1&60.5&67.4\\
     \textbf{CASD (reproduce)} & 68.8 & 67.2 & 53.9& 38.2& 21.5& 70.4& 69.7 & 68.9& 23.6 & 66.3 & 48.8 & 62.3 & 56.4 & 70.6 & 17.3 & 24.9 & 55.9 & 58.9 & 66.0 & 69.1 & 68.5\\
    \textbf{OICR+REG+W2N (Ours)} &87.4&86.0&69.7&50.8&59.8&89.8&88.4&86.9&37.5&86.5&26.0&69.8&84.0&95.1&31.6&57.6&78.12&75.6&85.8&77.3&\textbf{71.2(+3.8)}\\
    \textbf{CASD+W2N (Ours)}& 92.0&90.5&82.4&71.3&73.0&85.5&94.7&89.0&46.3&89.4&63.5&87.9&92.7&96.7&47.1&70.2&84.4&75.1&82.4&87.5&\textbf{80.1(+12.6)}\\
    \specialrule{.15em}{.05em}{.05em}
    \textbf{WSOD with transfer learning}:\\
    OICR+UBBR \cite{ubbr2018} & 47.9 & 18.9 & 63.1 & 39.7 & 10.2 & 62.3 & 69.3 & 61.0 & 27.0 & 79.0 & 24.5 & 67.9 & 79.1 & 49.7 & 28.6 & 12.8 & 79.4 & 40.6 & 61.6 & 28.4 & 47.6 \\
    WSLAT-Ens \cite{wslat2015} & 78.6 & 63.4 & 66.4 & 56.4 & 19.7 & 82.3 & 74.8 & 69.1 & 22.5 & 72.3 & 31.0 & 63.0 & 74.9 & 78.4 & 48.6 & 29.4 & 64.6 & 36.2 & 75.9 & 69.5  & 58.8 \\
    $\text{MSD-Ens}^{+}$ \cite{msd2018} & 89.2 & 75.7 & 75.1 & 66.5 & 58.8 & 78.2 & 88.9 & 66.9 & 28.2 & 86.3 & 29.7 & 83.5 & 83.3 & 92.8 & 23.7 & 40.3 & 85.6 & 48.9 & 70.3 & 68.1 & 66.8 \\
    $\text{Zhong \textit{et al.} (R50-C4)}^{+}$ \cite{zhong2020boosting} & 87.5 & 64.7 & 87.4 & 69.7 & 67.9 & 86.3 & 88.8 & 88.1 & 44.4 & 93.8 & 31.9 & 89.1 & 92.9 & 86.3 & 71.5 & 22.7 & 94.8 & 56.5 & 88.2 & 76.3 & 74.4 \\
    $\text{LBBA}^{+}$ \cite{dong2021boosting} & 93.3 & 90.6 & 71.8 & 69.2 & 59.5 & 90.9 & 94.4 & 78.5 & 55.4 & 96.6 & 51.0 & 82.3 & 72.5 & 93.2 & 48.5 & 52.8 & 100.0 & 66.7 & 78.3 & 87.5 & 76.7   \\
    $\text{TraMaS}^{+}$ \cite{2021arXiv211014191L}& 90.6 &67.4& 89.7 &70.5 &72.8 &86.6 &91.7 &89.8& 51.0 &96.1 &34.0 &93.7 &94.8 &90.3& 73.0& 26.5& 95.2 &68.2 &89.8 &83.1 & 77.7 \\
    $\text{CaT}_{5}$ \cite{Cao_2021_ICCV} &-&-&-&-&-&-&-&-&-&-&-&-&-&-&-&-&-&-&-&-& 80.3\\
     \textbf{LBBA (reproduce)} & 86.9&84.5&74.6&65.6&55.1&85.4&86.8&84.4&42.5&88.0&45.0&83.3&82.3&88,6&47.6&49.1&88.3&50.8&81.1&84.3&72.7\\
    \textbf{LBBA+W2N (Ours)}& 89.5&93.4&83.9&70.2&73.4&87.1&94.5&92.0&58.9&95.7&64.0&91.0&94.8&93.5&80.7&64.1&91.7&78.2&84.3&89.1&\textbf{83.5(+10.8)} \\
    \specialrule{.15em}{.05em}{.05em}
    \textbf{Upper bounds:} \\
    \hline
    Faster R-CNN (Res50+FPN)\cite{renNIPS15fasterrcnn} & 91.7 & 93.7 & 92.6 & 75.0 & 84.0 & 95.4 & 95.3 & 93.2 & 76.5 & 94.5 & 86.9 & 92.3 & 96.0 & 93.2 & 93.0 & 76.8 & 94.9 & 89.2 & 85.7 & 90.4 & \textbf{89.5} \\
    \specialrule{.15em}{.05em}{.05em}
    \end{tabular}
    }}
    \label{table:per-cls-voc07-corloc}
    \end{table*}

\section{Experiments}

\label{sec:exp}
\subsection{Experiment Settings}

\textbf{Datasets.}  Following~\cite{ren-wetectron2020,uijlings2018revisiting,zhong2020boosting}, we evaluate our method on four benchmarks: PASCAL VOC 2007, PASCAL VOC 2012~\cite{everingham2010pascal}, MS-COCO~\cite{lin2014microsoft}, and ILSVRC 2013~\cite{deng2009imagenet} detection dataset.
\noindent\textbf{Evaluation Metrics.} We use mean average precision (mAP) to evaluate the detection performance over categories, and CorLoc to measure the localization accuracy.

\subsection{Comparison with State-of-the-arts}
We \emph{state the implementation details in the suppl.}
And here we compare our method with several state-of-the-art WSOD approaches in terms of mAP and CorLoc on PASCAL VOC 2007~\cite{everingham2010pascal} reported by Table~\ref{table:per-cls-voc07} and Table~\ref{table:per-cls-voc07-corloc}. Our all results are obtained with single-scale testing approch. Based on these results, we obtain the following observations: First, our W2N framework outperforms all WSOD baselines in terms of both mAP and CorLoc. Specifically, on PASCAL VOC 2007 dataset, it outperforms OICR+REG by  8.7\% mAP and 3.8\% CorLoc, outperforms CASD by 11.4\% mAP and 12.6\% CorLoc, and outperforms LBBA by 9.5\% mAP and  10.8\% CorLoc. Performance on PASCAL VOC 2012 also demonstrates favorable performance improvement.

Second, our W2N outperforms all of the state-of-the-art WSOD methods as well as transfer learning based methods. Specifically, CASD+W2N achieves 65.4\% mAP on PASCAL VOC 2007 test set, which outperforms CASD by 8.6\% mAP and outperforms $\text{CaT}_{5}$ by 1.9\% mAP. Moreover, LBBA+W2N obtains 68.6\% mAP  and 83.4\% CorLoc, which achieves a new state-of-the-arts for WSOD problem and bridges the performance gap with fully supervised methods (Faster R-CNN)\cite{renNIPS15fasterrcnn}. In the supplementary we will show  more results on other datasets and give analyze for comparison between \cite{sui2021salvage} and ours.

\vspace{-1em}
\subsection{Ablation Study}

In this section, we discuss the effect of key components of W2N on PASCAL VOC 2007 dataset~\cite{everingham2010pascal}.

\begin{table}[t]

\parbox[t]{.45\textwidth}{
\caption{Effect of two modules on VOC 2007.}

  \begin{tabular}{l | c | c | c | c | c }
  
    \specialrule{.15em}{.05em}{.05em}
    WSOD & FRCNN* & LA & SSL & ITER & mAP \\
    \hline
    \multirow{5}{*}{LBBA} &  & & & & 59.1 \\
                          \cline{2-5}
                          & \checkmark & & & & 59.4 \\
                          &  & \checkmark & & & 60.3 \\
                          &  &  & \checkmark & & 66.1 \\
                          &  & \checkmark & \checkmark & & 67.0 \\
                          \cline{2-5}
                        &  & \checkmark & \checkmark &\checkmark & \textbf{68.6} \\
    
    \specialrule{.15em}{.05em}{.05em}
    \end{tabular}
  \label{table:com}
}
\hfill
\parbox[t]{.55\textwidth}{
\caption{The mAP results of our W2N with different iteration times $T$ on Pascal VOC 2007 dataset.}

    \begin{tabular}{l | c | c | c | c | c}
    \specialrule{.15em}{.05em}{.05em}
    Methods & 0 & 1  & 2 & 3 & 4 \\
    \hline
    OICR+REG+W2N & 56.8 & \textbf{57.0} & 56.8 & 56.8 & 56.9 \\
    CASD+W2N & 62.7 & 64.5 & \textbf{65.4} & 65.4 & 65.2 \\
    LBBA+W2N & 67 & 67.9 & \textbf{68.6} & 68.4 & 68.4\\
    \specialrule{.15em}{.05em}{.05em}
    \end{tabular}
    \centering

    \label{table:iter}
}

\end{table}

\begin{figure}[htbp]
\begin{minipage}[t]{0.47\linewidth}
\centering
    \includegraphics[scale=0.3]{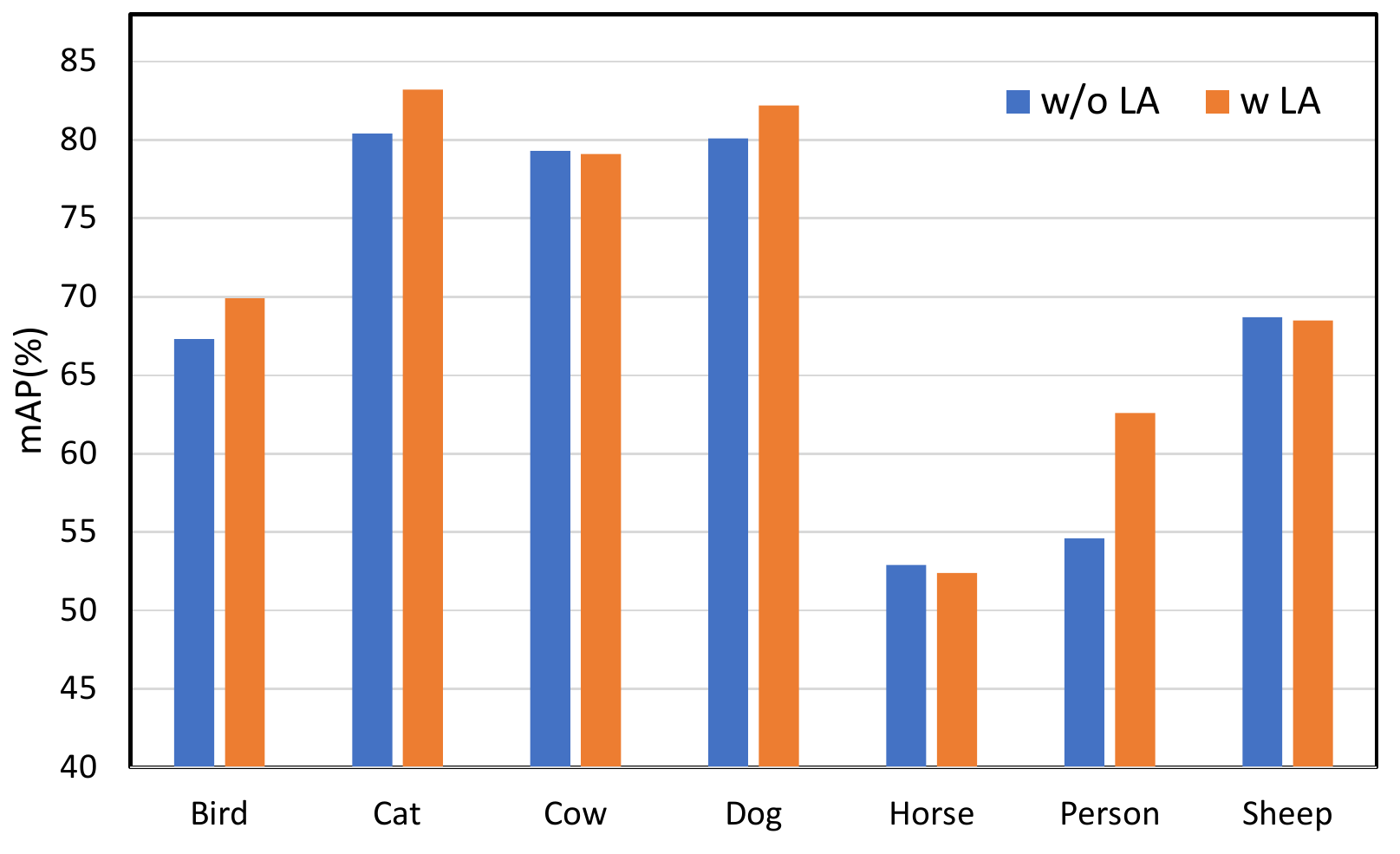}

      \caption{Effect of location adaption module on animal categories and person category with LBBA+W2N.}

    \label{fig:animal}

\end{minipage}
\hfill
\begin{minipage}[t]{0.47\linewidth}
\centering
    \includegraphics[scale=0.3]{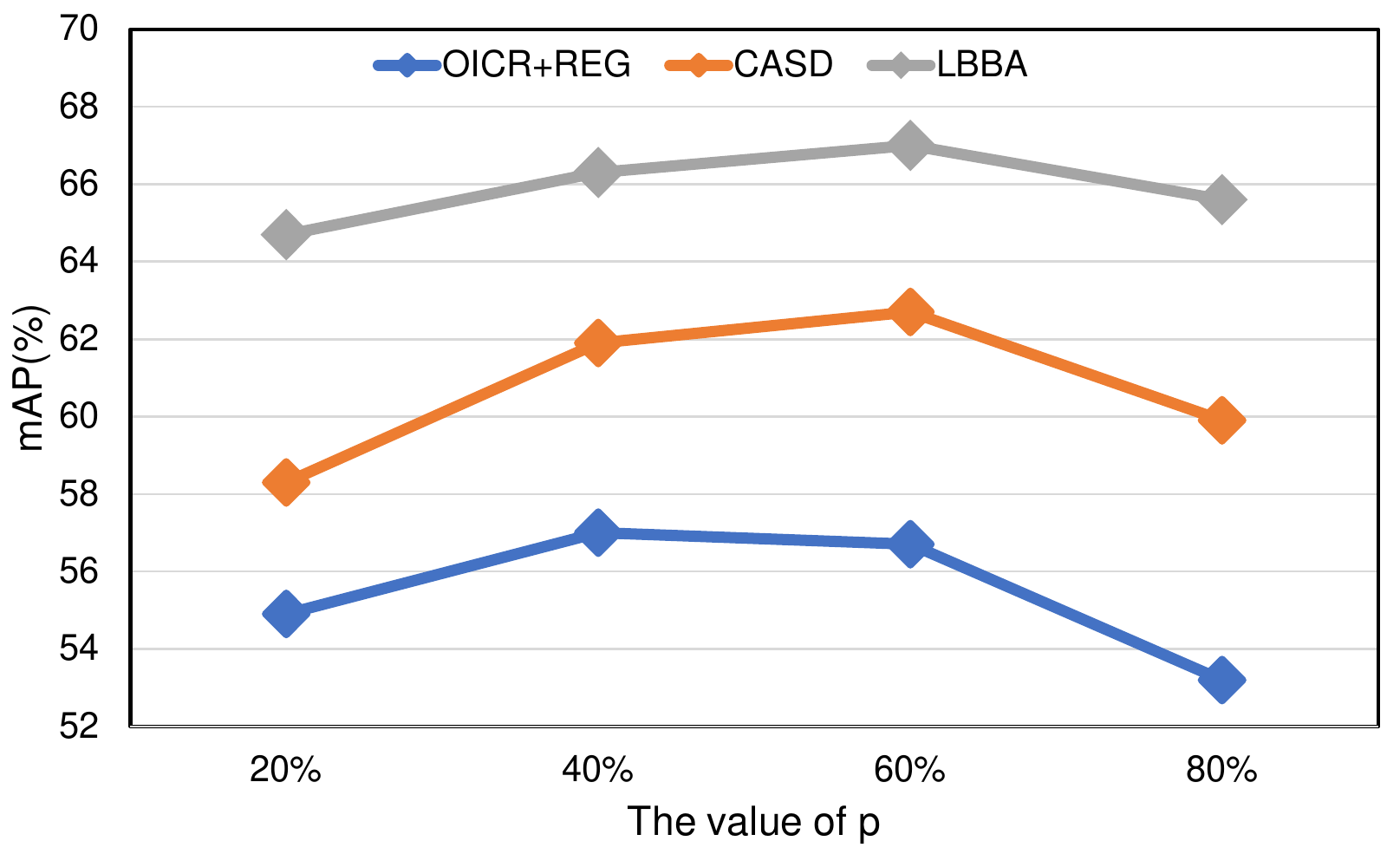}

      \caption{Effect of different size of the labeled set on VOC 2007 for different WSOD+W2Ns}

    \label{fig:rate}
\end{minipage}

\end{figure}

\noindent\textbf{Effect of Two Modules.} Table~\ref{table:com} shows the ablation study of each module on LBBA
baseline.
Simply re-training Faster R-CNN(FRCNN*) with pseudo GT only brings 0.3\% mAP gain. 
By introducing localization adaption and semi-supervised learning separately, these improvements respectively outperform the baseline by 1.2\% and 7.0\% in terms of mAP. 
Specifically, as illustrated in Fig.~\ref{fig:animal}, W2N+LBBA with location adaption module improves the detection performance of categories which suffer from the discriminative part problem, especially for person category.
Furthermore, our full method combining these two modules can further improve the detection performance to 67.0\% mAP. 
More ablation study about effect of two modules can be found in the suppl.

\noindent\textbf{Effect of Iterative Training. } 
Generally, more training iterations means better predictions. 
Thus we analyze the effect of training iteration $T$. Table.~\ref{table:iter} shows the performance of W2N with different iteration numbers $T$ using three different methods, respectively. Generally, as the $T$ increases, the performance first increase and then begin to oscillate near the highest point. And the highest performance for all baseline are outperforming beyond 1.5\% mAP than the settings of $T=0$, which proves that the iterative training strategy is effective for further improving detection performance. In addition, for LBBA and CASD, it reaches the highest performance when $T=2$; while for OICR+REG, $T=1$ is the best optimal solution. This result indicates that the iterative training process will converge quickly on relative small $T$, which reveals the high efficiency of W2N.
    
\begin{table}[t]

    \caption{Comparisons of different dataset split methods on VOC 2007. }

    \centering
    \begin{tabular}{l | c | c | c | c }
    \specialrule{.15em}{.05em}{.05em}
    Methods & image-level & instance-level & two tasks instance-level & mAP\\
    \hline
    OICR+REG+W2N & \checkmark &  &  & 55.4 \\
    OICR+REG+W2N &  & \checkmark & & 56.8 \\
    OICR+REG+W2N &  & & \checkmark & \textbf{56.8} \\
    \hline
    CASD+W2N & \checkmark &  & & 61.9\\
    CASD+W2N &  & \checkmark & & 62.6 \\
    CASD+W2N &  & & \checkmark & \textbf{62.7} \\
    \hline
    LBBA+W2N & \checkmark &  &  & 65.4\\
    LBBA+W2N &  & \checkmark & & 66.8 \\
    LBBA+W2N &  &  & \checkmark & \textbf{67.0}\\
    \specialrule{.15em}{.05em}{.05em}
    \end{tabular}
    \label{table:split}

    \end{table}
    
\noindent\textbf{Effect of Hybrid-Level Dataset Split. } 
we combined three different WSOD methods with three different split methods and then obtained nine different experiment settings. We conducted experiments on all of the settings at iteration 0 and demonstrate the results in Table.~\ref{table:split}. Experimental results prove that the two tasks instance-level split method achieves the best performance among them, higher than the instance-level split method. In addition, both two tasks instance-level split method and instance-level split method outperform the image-level split method more than about 1.5\% mAP, which proves that it is more effective and reasonable to treat the instance-level as the smallest division unit. 

\noindent\textbf{Proportion of Clean Split $p$. } 
The proportion of clean split $p$ determines the quality of pseudo labels, therefore here we explore the effect of different $p$.  We deploy varying $p$ to decide the size of labeled set for three different WSOD methods at iteration 0.  Fig.~\ref{fig:rate} shows that for LBBA and CASD, $p=60\%$ is the best choice, while for OICR+REG, $p=40\%$ is better. Generally, when $p$ is small, as $p$ increases, the performance of W2Ns improves, while $p$ further increases, the performance of W2Ns begin to drop significantly. This is reasonable that too small leads to a small size of high quality pseudo label in labeled set, which is not conducive to model learning. While too large clean size will involve more noisy labels. Therefore, we propose that a moderate size is beneficial for training.






\section{Conclusion}
In this paper, we propose a weakly supervised object detection method namely Weakly-supervision to Noisy-supervision (W2N). We treat the pseudo labels generated by the pretrained weakly detector as noisy labels and propose an iterative training procedure, which includes the localization adaptation module and the semi-supervised learning module. The localization adaptation module refines the original pseudo ground-truths to reduce the proportion of low-quality bounding boxes. The semi-supervised learning module split the dataset with pseudo ground-truths into a high-quality labeled set as well as an unlabeled set and supervises the object detector with a well-designed semi-supervised object detection manner with these two datasets. Extensive experiments on different datasets show that our proposed method performs favorably against other state-of-the-art WSOD methods.

\subsubsection{Acknowledgement}
This work was supported in part by the National Key R$\&$D Program of China under Grant No. 2021ZD0112100, and the Major Key Project of PCL under Grant No. PCL2021A12. This work was done when Zitong was an intern at MEGVII Tech.

\clearpage

%
%
\bibliographystyle{splncs04}
\bibliography{egbib}
\clearpage
\appendix
\begin{center}
  \Large
  \textbf{Appendix}
\end{center}

\begin{table*}[h!]
    \caption{Comparison of our method on PASCAL VOC 2007 test set to  state-of-the-art WSOD methods in terms of mAP ($\%$), where $\text{~}^{+}$ means the results with multi-scale testing.}
    \centering
    \resizebox{\textwidth}{!}{
    {
    \begin{tabular}{l | c c c c c c c c c c c c c c c c c c c c | c}
    \specialrule{.15em}{.05em}{.05em}
    Methods & Aero & Bike & Bird & Boat & Bottle & Bus & Car & Cat & Chair & Cow & Table & Dog & Horse & Motor & Person & Plant & Sheep & Sofa & Train & ~~TV~~ & ~~AP~~ \\
    \hline
    \textbf{Pure WSOD}:\\
    WSDDN~\cite{bilen2016weakly} &39.4 & 50.1 & 31.5 & 16.3 & 12.6 & 64.5 & 42.8 & 42.6 & 10.1 & 35.7 & 24.9 & 38.2 & 34.4 & 55.6 & 9.4 & 14.7 & 30.2 & 40.7 & 54.7 & 46.9 & 34.8 \\
    $\text{OICR}^{+}$ \cite{tang2017multiple} & 58.0 & 62.4 & 31.1 & 19.4 & 13.0 & 65.1 & 62.2 & 28.4 & 24.8 & 44.7 & 30.6 & 25.3 & 37.8 & 65.5 & 15.7 & 24.1 & 41.7 & 46.9 & 64.3 & 62.6 & 41.2 \\
    $\text{PCL}^{+}$ \cite{tang2018pcl} & 54.4 & 69.0 & 39.3 & 19.2 & 15.7 & 62.9 & 64.4 & 30.0 & 25.1 & 52.5 & 44.4 & 19.6 & 39.3 & 67.7 & 17.8 & 22.9 & 46.6 & 57.5 & 58.6 & 63.0 & 43.5 \\
    $\text{Yang }\textit{et al.}^{+}$ \cite{yang2019towards} & 57.6 & 70.8 & 50.7 & 28.3 & 27.2 & 72.5 & 69.1 & 65.0 & 26.9 & 64.5 & 47.4 & 47.7  & 53.5 & 66.9 & 13.7 & 29.3 & 56.0 & 54.9 & 63.4 & 65.2 & 51.5 \\
    $\text{C-MIDN}^{+}$ \cite{Gao_2019_ICCV} & 53.3 & 71.5 & 49.8 & 26.1 & 20.3 & 70.3 & 69.9 & 68.3 & 28.7 & 65.3 & 45.1 & {64.6} & 58.0 & 71.2 & 20.0 & 27.5 & 54.9 & 54.9 & {69.4} & 63.5 & 52.6 \\
    Arun \textit{et al.} \cite{Arun_2019} & 66.7 & 69.5 & 52.8 & 31.4 & 24.7 & {74.5} & 74.1 & 67.3 & 14.6 & 53.0 & 46.1 & 52.9 & 69.9 & 70.8 & 18.5 & 28.4 & 54.6 & {60.7} & 67.1 & 60.4 & 52.9 \\
    $\text{WSOD2}^{+}$ \cite{Zeng_2019_ICCV} & 65.1 & 64.8 & {57.2} & {39.2} & 24.3 & 69.8 & 66.2 & 61.0 & 29.8 & 64.6 & 42.5 & 60.1 & {71.2} & 70.7 & 21.9 & 28.1 & 58.6 & 59.7 & 52.2 & 64.8 & 53.6 \\
    GradingNet-C-MIL \cite{Jia_Wei_Ruan_Zhao_Zhao_2021} &-&-&-&-&-&-&-&-&-&-&-&-&-&-&-&-&-&-&-&-& 54.3 \\
    MIST-Full \cite{ren-wetectron2020} & {68.8} & {77.7} & 57.0 & 27.7 & {28.9} & 69.1 & {74.5} & 67.0 & {32.1} & {73.2} & 48.1 & 45.2 & 54.4 & {73.7} & {35.0} & 29.3 & {64.1} & 53.8 & 65.3 & {65.2} & {54.9}  \\
    $\text{IM-CFB}^{+}$ \cite{Yin_Deng_Zhou_Li_2021} & 63.3 & 77.5 & 48.3 & 36.0 & 32.6 & 70.8 & 71.9 & 73.1 & 29.1 & 68.7 & 47.1 & 69.4 & 56.6 & 70.9 & 22.8 & 24.8 & 56.0 & 59.8 & 73.2 & 64.6 & 55.8 \\
    CASD \cite{huangCASD2020} &-&-&-&-&-&-&-&-&-&-&-&-&-&-&-&-&-&-&-&-&  56.8 \\
    SoS \cite{sui2021salvage} & 72.9 & 79.4& 59.6 &20.4 &49.8 &81.2 &82.9 &84.0 &31.5& 76.6& 57.4 &60.7& 74.7 &75.1& 33.0& 34.3 &66.3 &61.1& 80.6& 71.8 & 62.7\\
    $\text{SoS}^{+}$ \cite{sui2021salvage} & 77.9 &  81.2  & 58.9  & 26.7 &  54.3 &  82.5  & 84.0 &  83.5 & 36.3 &  76.5  & 57.5 &  58.4  & 78.5  & 78.6  & 33.8  & 37.4  & 64.0 &  63.4 &  81.5 &  74.0 & 64.4 \\

     \textbf{OICR+REG (reproduce)} & 54.0&61.9&43.9&22.6&31.7&73.8&65.1&60.6&14.4&68.0&17.0&48.8&58.3&69.9&12.8&22.0&53.9&53.6&69.7&60.4&48.3\\
    \textbf{CASD (reproduce)} & 68.8 & 67.2 & 53.9& 38.2& 21.5& 70.4& 69.7 & 68.9& 23.6 & 66.3 & 48.8 & 62.3 & 56.4 & 70.6 & 17.3 & 24.9 & 55.9 & 58.9 & 66.0 & 69.1 & 54.0\\
  
    \textbf{OICR+REG+W2N (Ours)} &71.0&74.2&60.8&28.8&44.6&78.0&72.6&80.3&16.7&74.3&24.3&58.2&64.6&75.1&13.3&29.9&60.3&65.3&80.1&67.6&\textbf{57.0(+8.7)}\\
    \textbf{CASD+W2N (Ours)}& 74.0&81.7&71.2&48.9&51.0&78.6&82.3&83.5&29.1&76.9&51.5&82.1&76.9&79.1&28.5&34.3&65.0&64.2&75.2&74.8&\textbf{65.4(+11.4)}\\
    \specialrule{.15em}{.05em}{.05em}
    \textbf{WSOD with transfer learning}:\\
    $\text{MSD-Ens}^{+}$ \cite{msd2018} & 70.5 & 69.2 & 53.3 & 43.7 & 25.4 & 68.9 & 68.7 & 56.9 & 18.4 & 64.2 & 15.3 & 72.0 & 74.4 & 65.2 & 15.4 & 25.1 & 53.6 & 54.4 & 45.6 & 61.4 & 51.1 \\
    OICR+UBBR \cite{ubbr2018} & 59.7 & 44.8 & 54.0 & 36.1 & 29.3 & 72.1 & 67.4 & 70.7 & 23.5 & 63.8 & 31.5 & 61.5 & 63.7 & 61.9 & 37.9 & 15.4 & 55.1 & 57.4 & 69.9 & 63.6 & 52.0 \\
    $\text{LBBA}^{+}$ \cite{dong2021boosting} & 70.3 & 72.3 & 48.7 & 38.7 & 30.4 & 74.3 & 76.6 & 69.1 & 33.4 & 68.2 & 50.5 & 67.0 & 49.0 & 73.6 & 24.5 & 27.4 & 63.1 & 58.9 & 66.0 & 69.2 & 56.6 \\
    $\text{Zhong \textit{et al.} (R50-C4)}^{+}$ \cite{zhong2020boosting} & 64.8 & 50.7 & 65.5 & 45.3 & 46.4 & 75.7 & 74.0 & 80.1 & 31.3 & 77.0 & 26.2 & 79.3 & 74.8 & 66.5 & 57.9 & 11.5 & 68.2 & 59.0 & 74.7 & 65.5 & 59.7 \\
    $\text{TraMaS}^{+}$ \cite{2021arXiv211014191L} & 68.6 & 61.1& 69.6& 48.1 &49.9 &76.3 &77.8& 80.9 &34.9& 77.0& 31.1& 80.9& 78.5& 66.3& 64.0 &19.1 &69.1& 62.3& 74.4& 69.1& 62.9\\
    $\text{CaT}_{5}$ \cite{Cao_2021_ICCV} & 74.0& 70.7& 60.0 &31.1& 50.0& 75.9& 82.0& 70.7& 32.8& 74.3& 69.5& 70.2 &69.5& 77.0& 37.5& 45.8& 67.0 &61.1& 72.4 &68.0 & 63.0\\
     \textbf{LBBA (reproduce)} & 70.2&75.5&49.2&41.9&30.5&80.5&78.2&72.8&36.4&73.8&52.3&67.0&46.4&76.2&34.6&29.4&67.9&66.6&68.3&74.1&59.1\\
    \textbf{LBBA+W2N (Ours)}& 71.8&83.0&69.9&50.3&54.5&79.0&83.9&83.9&39.4&79.2&52.9&82.2&83.6&79.2&62.6&32.7&68.5&66.1&75.8&74.5&\textbf{68.6(+9.5)} \\
    \specialrule{.15em}{.05em}{.05em}
    \textbf{Upper bounds:} \\
    \hline
    Faster R-CNN (Res50+FPN) \cite{renNIPS15fasterrcnn} &82.8&84.2&75.2&62.4&67.0&81.4&87.1&82.6&57.3&82.5&64.9&83.0&84.0&82.7&83.7&54.0&76.1&73.4&81.8&76.1&\textbf{76.1}\\

    \specialrule{.15em}{.05em}{.05em}
    \end{tabular}
    }}
    \label{table:per-cls-voc07}
  
\end{table*}

\section{Experiments}
\subsection{Implementation Details}
In this subsection we show some implementation details of experiments.

\paragraph{Overall.} All programs are conducted based on  PyTorch toolkit and run on NVIDIA GTX1080Ti GPU $\times 8$. 

\paragraph{Training of Weakly Supervised Detector.} In this work we select three WSOD baseline methods to play the role of generators: \emph{OICR+REG}, \emph{CASD} and \emph{LBBA}. For OICR+REG, we adopt the code by \cite{DRN-WSOD_2020_ECCV}, and for CASD and LBBA, we adopt code with official implementations. To make a long story short, the is no any trick introduced in training phase and all the training configuration and hyper-parameter are the same as the default version of the code base, except for that we select the COCO-60-clean \cite{dong2021boosting,zhong2020boosting} as the auxiliary dataset to train the LBBA \cite{dong2021boosting} network. 

\paragraph{Noisy Label Generation.} \label{sec:nlg} We follow  the Pseudo Ground-truth Excavation (PGE) algorithm \cite{zhang2018w2f} to implement the noisy label generation on training set, where the threshold $T_{nms}$ for NMS is set to 0.3, while $T_{score}$ and $T_{fusion}$ are set to 0.2 and 0.4 respectively. 

\paragraph{Learning Detector with Noisy Annotations}
For localization adaptation stage, we adopt Faster R-CNN \cite{renNIPS15fasterrcnn} with backbone of ResNet-50 \cite{He_2016_CVPR} combined with FPN \cite{Lin_2017_CVPR} as the supervised object object detector $f$ .During training, the $f$ is optimized by stochastic gradient descent (SGD) \cite{bottou2010large} with the batch size of 16, initialized learning rate of 0.02, momentum of 0.9 and weight decay of $1 \times 10^{-4}$. The number steps of training is set to 5,000, 10,000, 90,000, 250,000 and the learning rate is decayed by 0.1 after 3,500, 7,000, 60,000, 180,000 steps for PASCAL VOC 2007, PASCAL VOC 2012, MS-COCO and ILSVRC respectively. $\tau_{score}$, $\tau_{assign}$, $\lambda_{re}$, $\alpha$ and $\beta$ are set to 0.1, 0.5, 0.1, 0.05 and 0.8 respectively. After training, we use $f$ to refine the noisy labels set $\mathbb{X}_{p}$ by the same procedure of the noisy label generation. For semi-supervised learning stage, we choose the \emph{two tasks instance-level} data split method and the proportion $p$ of clean data is set to 60\%. We implement the semi-supervised object detection (SSOD) framework by Unbiased Teacher \cite{liu2020unbiased} with its official code. During SSOD training, batch size of labeled set and unlabeled set are set to 8, and the learning rate is set to 0.01, the number of training step is set to 30,000, 50,000, 100,000 and 200,000 for PASCAL VOC 2007, PASCAL VOC 2012, MS-COCO and ILSVRC respectively and learning rate decay is not adopted. The iteration time $T$ for the whole iterative manner is set to 2 for CASD and LBBA while set to 1 for OICR+REG and the further ablation study has shown in \emph{submission file}. The weight of unsupervised loss $\lambda_{u}$ is set to 2 for all settings. Other hyper-parameters are adopted the default configuration of \cite{liu2020unbiased}.

\begin{table*}[t!]
    \caption{Comparison of our method on PASCAL VOC 2007 trainval set to  state-of-the-art WSOD methods in terms of CorLoc ($\%$), where $\text{~}^{+}$ means the results with multi-scale testing.}
    \centering
  
    \resizebox{\textwidth}{!}{
    {
    \begin{tabular}{l | c c c c c c c c c c c c c c c c c c c c | c}
    \specialrule{.15em}{.05em}{.05em}
    Methods & Aero & Bike & Bird & Boat & Bottle & Bus & Car & Cat & Chair & Cow & Table & Dog & Horse & Motor & Person & Plant & Sheep & Sofa & Train & ~~TV~~ & CorLoc \\
    \hline
    \textbf{Pure WSOD}: \\
    WSDDN~\cite{bilen2016weakly} & 65.1 & 58.8 & 58.5 & 33.1 & 39.8 & 68.3 & 60.2 & 59.6 & 34.8 & 64.5 & 30.5 & 43.0 & 56.8 & 82.4 & 25.5 & 41.6 & 61.5 & 55.9 & 65.9 & 63.7 & 53.5 \\
    $\text{OICR}^{+}$ \cite{tang2017multiple} & 81.7 & 80.4 & 48.7 & 49.5 & 32.8 & 81.7 & 85.4 & 40.1 & 40.6 & 79.5 & 35.7 & 33.7 & 60.5 & 88.8 & 21.8 & 57.9 & 76.3 & 59.9 & 75.3 & {81.4} & 60.6 \\
    $\text{PCL}^{+}$ \cite{tang2018pcl} & 79.6 & 85.5 & 62.2 & 47.9 & 37.0 & 83.8 & 83.4 & 43.0 & 38.3 & 80.1 & 50.6 & 30.9 & 57.8 & 90.8 & 27.0 & 58.2 & 75.3 & {68.5} & 75.7 & 78.9 & 62.7 \\
    $\text{Li}^{+}$ \cite{li2019weakly} & 85.0 & 83.9 & 58.9 & 59.6 & 43.1 & 79.7 & 85.2 & 77.9 & 31.3 & 78.1 & 50.6 & 75.6 & 76.2 & 88.4 & 49.7 & 56.4 & 73.2 & 62.6 & 77.2 & 79.9 & 68.6 \\
    $\text{C-MIL}^{+}$ \cite{Wan_2019} &-&-&-&-&-&-&-&-&-&-&-&-&-&-&-&-&-&-&-&-& 65.0 \\
    $\text{Yang }\textit{et al.}^{+}$ \cite{yang2019towards} & 80.0 & 83.9 & 74.2 & 53.2 & 48.5 & 82.7 & 86.2 & 69.5 & 39.3 & 82.9 & 53.6 & 61.4& 72.4 & 91.2 & 22.4 & 57.5 & {83.5} & 64.8 & 75.7 & 77.1 & 68.0 \\
    $\text{MIST (Full)}^{+}$ \cite{ren-wetectron2020} & 87.5 & 82.4 & {76.0} & 58.0 & 44.7 & 82.2 & 87.5 & 71.2 & {49.1} & 81.5 & 51.7 & 53.3 & 71.4 & 92.8 & 38.2 & 52.8 &79.4 & 61.0 & 78.3 & 76.0  & 68.8 \\
    $\text{WSOD2}^{+}$ \cite{Zeng_2019_ICCV} & 87.1 & 80.0 & 74.8 & 60.1 & 36.6 & 79.2 & 83.8 & 70.6 & 43.5 & {88.4} & 46.0 & {74.7} & 87.4 & 90.8 & 44.2 & 52.4 & 81.4 & 61.8 & 67.7 & 79.9 & 69.5\\
    Arun~ \textit{et al.}\cite{Arun_2019} & {88.6} & {86.3} & 71.8 & 53.4 & 51.2 & {87.6} & {89.0} & 65.3 & 33.2 & 86.6 & 58.8 & 65.9 & {87.7} & {93.3} & 30.9 & 58.9 & 83.4 & 67.8 & 78.7 & 80.2 & {70.9} \\
    GradingNet-C-MIL \cite{Jia_Wei_Ruan_Zhao_Zhao_2021} &-&-&-&-&-&-&-&-&-&-&-&-&-&-&-&-&-&-&-&-&  72.1 \\
    $\text{IM-CFB}^{+}$ \cite{Yin_Deng_Zhou_Li_2021} &-&-&-&-&-&-&-&-&-&-&-&-&-&-&-&-&-&-&-&-& 72.2 \\
     \textbf{OICR+REG (reproduce)} & 91.6&78.3&62.6&46.0&44.8&86.4&87.7&80.3&34.4&87.1&30.1&69.4&81.1&90.8&31.3&44.8&76.0&76.1&83.1&60.5&67.4\\
     \textbf{CASD (reproduce)} & 68.8 & 67.2 & 53.9& 38.2& 21.5& 70.4& 69.7 & 68.9& 23.6 & 66.3 & 48.8 & 62.3 & 56.4 & 70.6 & 17.3 & 24.9 & 55.9 & 58.9 & 66.0 & 69.1 & 68.5\\
    \textbf{OICR+REG+W2N (Ours)} &87.4&86.0&69.7&50.8&59.8&89.8&88.4&86.9&37.5&86.5&26.0&69.8&84.0&95.1&31.6&57.6&78.12&75.6&85.8&77.3&\textbf{71.2(+3.8)}\\
    \textbf{CASD+W2N (Ours)}& 92.0&90.5&82.4&71.3&73.0&85.5&94.7&89.0&46.3&89.4&63.5&87.9&92.7&96.7&47.1&70.2&84.4&75.1&82.4&87.5&\textbf{80.1(+12.6)}\\
    \specialrule{.15em}{.05em}{.05em}
    \textbf{WSOD with transfer learning}:\\
    OICR+UBBR \cite{ubbr2018} & 47.9 & 18.9 & 63.1 & 39.7 & 10.2 & 62.3 & 69.3 & 61.0 & 27.0 & 79.0 & 24.5 & 67.9 & 79.1 & 49.7 & 28.6 & 12.8 & 79.4 & 40.6 & 61.6 & 28.4 & 47.6 \\
    WSLAT-Ens \cite{wslat2015} & 78.6 & 63.4 & 66.4 & 56.4 & 19.7 & 82.3 & 74.8 & 69.1 & 22.5 & 72.3 & 31.0 & 63.0 & 74.9 & 78.4 & 48.6 & 29.4 & 64.6 & 36.2 & 75.9 & 69.5  & 58.8 \\
    $\text{MSD-Ens}^{+}$ \cite{msd2018} & 89.2 & 75.7 & 75.1 & 66.5 & 58.8 & 78.2 & 88.9 & 66.9 & 28.2 & 86.3 & 29.7 & 83.5 & 83.3 & 92.8 & 23.7 & 40.3 & 85.6 & 48.9 & 70.3 & 68.1 & 66.8 \\
    $\text{Zhong \textit{et al.} (R50-C4)}^{+}$ \cite{zhong2020boosting} & 87.5 & 64.7 & 87.4 & 69.7 & 67.9 & 86.3 & 88.8 & 88.1 & 44.4 & 93.8 & 31.9 & 89.1 & 92.9 & 86.3 & 71.5 & 22.7 & 94.8 & 56.5 & 88.2 & 76.3 & 74.4 \\
    $\text{LBBA}^{+}$ \cite{dong2021boosting} & 93.3 & 90.6 & 71.8 & 69.2 & 59.5 & 90.9 & 94.4 & 78.5 & 55.4 & 96.6 & 51.0 & 82.3 & 72.5 & 93.2 & 48.5 & 52.8 & 100.0 & 66.7 & 78.3 & 87.5 & 76.7   \\
    $\text{TraMaS}^{+}$ \cite{2021arXiv211014191L}& 90.6 &67.4& 89.7 &70.5 &72.8 &86.6 &91.7 &89.8& 51.0 &96.1 &34.0 &93.7 &94.8 &90.3& 73.0& 26.5& 95.2 &68.2 &89.8 &83.1 & 77.7 \\
    $\text{CaT}_{5}$ \cite{Cao_2021_ICCV} &-&-&-&-&-&-&-&-&-&-&-&-&-&-&-&-&-&-&-&-& 80.3\\
     \textbf{LBBA (reproduce)} & 86.9&84.5&74.6&65.6&55.1&85.4&86.8&84.4&42.5&88.0&45.0&83.3&82.3&88,6&47.6&49.1&88.3&50.8&81.1&84.3&72.7\\
    \textbf{LBBA+W2N (Ours)}& 89.5&93.4&83.9&70.2&73.4&87.1&94.5&92.0&58.9&95.7&64.0&91.0&94.8&93.5&80.7&64.1&91.7&78.2&84.3&89.1&\textbf{83.5(+10.8)} \\
    \specialrule{.15em}{.05em}{.05em}
    \textbf{Upper bounds:} \\
    \hline
    Faster R-CNN (Res50+FPN)\cite{renNIPS15fasterrcnn} & 91.7 & 93.7 & 92.6 & 75.0 & 84.0 & 95.4 & 95.3 & 93.2 & 76.5 & 94.5 & 86.9 & 92.3 & 96.0 & 93.2 & 93.0 & 76.8 & 94.9 & 89.2 & 85.7 & 90.4 & \textbf{89.5} \\
    \specialrule{.15em}{.05em}{.05em}
    \end{tabular}
    }}

    \label{table:per-cls-voc07-corloc}
    \end{table*}

\begin{table*}[t]
    \caption{Comparison of our method on PASCAL VOC 2012 test set to  state-of-the-art WSOD methods in terms of mAP ($\%$), where $\text{~}^{+}$ means the results with multi-scale testing.}
    \centering
    \resizebox{\textwidth}{!}{
    {
    \begin{tabular}{l | c c c c c c c c c c c c c c c c c c c c | c}
    \specialrule{.15em}{.05em}{.05em}
    Methods & Aero & Bike & Bird & Boat & Bottle & Bus & Car & Cat & Chair & Cow & Table & Dog & Horse & Motor & Person & Plant & Sheep & Sofa & Train & ~~TV~~ & ~~AP~~ \\
    \hline
    \textbf{Pure WSOD}: \\
    $\text{OICR}^{+}$ \cite{tang2017multiple} & 67.7 & 61.2 & 41.5 & 25.6 & 22.2 & 54.6 & 49.7 & 25.4 & 19.9 & 47.0 & 18.1 & 26.0 & 38.9 & 67.7 & 2.0 & 22.6 & 41.1 & 34.3 & 37.9 & 55.3 & 37.9 \\
    $\text{PCL}^{+}$ \cite{tang2018pcl} & 58.2 & 66.0 & 41.8 & 24.8 & 27.2 & 55.7 & 55.2 & 28.5 & 16.6 & 51.0 & 17.5 & 28.6 & 49.7 & 70.5 & 7.1 & 25.7 & 47.5 & 36.6 & 44.1 & 59.2 & 40.6\\
    $\text{Yang }\textit{et al.}^{+}$ & 64.7 & 66.3 & 46.8 & 28.5 & 28.4 & 59.8 & 58.6 & 70.9 & 13.8 & 55.0 & 15.7 & 60.5 &  {63.9} & 69.2 & 8.7 & 23.8 & 44.7 &  {52.7} & 41.5 & 62.6 & 46.8 \\
    $\text{WSOD2}^{+}$ \cite{Zeng_2019_ICCV} &-&-&-&-&-&-&-&-&-&-&-&-&-&-&-&-&-&-&-&-& 47.2\\
    Arun \textit{et al.} \cite{Arun_2019} &-&-&-&-&-&-&-&-&-&-&-&-&-&-&-&-&-&-&-&-& 48.4 \\
    IM-CFB \cite{Yin_Deng_Zhou_Li_2021} &-&-&-&-&-&-&-&-&-&-&-&-&-&-&-&-&-&-&-&-&49.4\\
    $\text{C-MIDN}^{+}$ \cite{Gao_2019_ICCV} & 72.9 & 68.9 & 53.9 & 25.3 & 29.7 & 60.9 & 56.0 &  {78.3} & 23.0 & 57.8 & 25.7 &  {73.0} & 63.5 & 73.7 & 13.1 & 28.7 & 51.5 & 35.0 & 56.1 & 57.5 & 50.2 \\
    GradingNet-C-MIL \cite{Jia_Wei_Ruan_Zhao_Zhao_2021} &-&-&-&-&-&-&-&-&-&-&-&-&-&-&-&-&-&-&-&-&50.5 \\
    $\text{MIST (Full)}^{+}$ \cite{ren-wetectron2020}  & {78.3} & {73.9} & {56.5} & {30.4} & {37.4} & {64.2} & {59.3} & 60.3 & {26.6} & {66.8} & 25.0 & 55.0 & 61.8 & {79.3} & 14.5 & {30.3} & {61.5} & 40.7 & {56.4} & {63.5} & {52.1}   \\
    CASD \cite{huangCASD2020} &-&-&-&-&-&-&-&-&-&-&-&-&-&-&-&-&-&-&-&-& 53.6 \\
    SoS \cite{sui2021salvage} &-&-&-&-&-&-&-&-&-&-&-&-&-&-&-&-&-&-&-&-& 59.6 \\
    $\text{SoS}^{+}$ \cite{sui2021salvage} &-&-&-&-&-&-&-&-&-&-&-&-&-&-&-&-&-&-&-&-& 61.9 \\
    \textbf{OICR+REG (reproduce)} &57.1&60.7&45.7&24.9&30.6&62.7&52.4&65.9&24.4&52.5&19.1&50.0&55.6&69.5&10.3&22.9&51.0&42.3&61.8&54.6&45.7\\ 
    \textbf{CASD (reproduce)} &65.9&69.3&55.3&27.9&40.2&61.6&61.6&75.3&32.4&55.2&22.7&51.6&60.0&74.7&10.0&27.1&55.1&48.2&68.3&61.1&51.2\\ 
    \textbf{OICR+REG+W2N (Ours)} &75.2&76.7&63.8&32.9&48.7&70.3&70.3&81.1&38.5&63.9&23.8&57.5&69.1&78.6&9.8&36.4&65.6&54.7&77.8&64.0&\textbf{57.9(+12.2)}\\
    \textbf{CASD+W2N (Ours)}& 81.8&78.9&69.8&33.5&48.0&75.0&73.9&84.9&33.2&71.1&16.2&84.9&78.1&78.4&11.2&38.9&71.7&45.7&77.5&64.5&\textbf{60.8(+9.6)}\\
    \hline
    \textbf{WSOD with transfer learning}:\\
    $\text{MSD-Ens}^{+}$ \cite{msd2018}  &-&-&-&-&-&-&-&-&-&-&-&-&-&-&-&-&-&-&-&-& 43.4 \\
    LBBA & {77.0} & {71.0} & {62.0} & {40.0} & {37.5} & {67.4} & {62.5} & 68.3 & {23.6} & {71.4} & 25.6 & 78.4 & 71.9 & {74.3} & 6.7 & {29.2} & {62.8} & 50.6 & {47.8} & {62.1} & \textbf{54.5}   \\
    \textbf{LBBA (reproduce)} &76.8&71.4&61.4&40.0&38.1&66.6&63.8&69.9&22.6&65.7&23.9&77.5&72.8&74.4&6.4&29.3&59.3&51.4&47.2&62.8&54.0\\
    \textbf{LBBA+W2N (Ours)}& 81.4&80.7&72.6&39.5&52.7&78.2&76.7&82.3&34.9&77.4&20.9&83.6&79.4&81.9&11.1&37.7&75.3&49.4&74.2&65.4&\textbf{62.7(+8.7)} \\

    \specialrule{.15em}{.05em}{.05em}
    \end{tabular}
    }}
    \label{table:per-cls-voc12}
    \end{table*}
    
\begin{table*}[tbh]
        \caption{Comparison of our method on PASCAL VOC 2012 trainval set to  state-of-the-art WSOD methods in terms of CorLoc ($\%$), where $\text{~}^{+}$ means the results with multi-scale testing.}
    \centering
    \resizebox{\textwidth}{!}{
    {
    \begin{tabular}{l | c c c c c c c c c c c c c c c c c c c c | c}
    \specialrule{.15em}{.05em}{.05em}
    Methods & Aero & Bike & Bird & Boat & Bottle & Bus & Car & Cat & Chair & Cow & Table & Dog & Horse & Motor & Person & Plant & Sheep & Sofa & Train & ~~TV~~ & CorLoc \\
    \hline
    \textbf{Pure WSOD}: \\
    $\text{OICR}^{+}$ \cite{tang2017multiple} &-&-&-&-&-&-&-&-&-&-&-&-&-&-&-&-&-&-&-&-& 62.1 \\
    $\text{PCL}^{+}$ \cite{tang2018pcl} & 77.2 & 83.0 & 62.1 & 55.0 & 49.3 & 83.0 & 75.8 & 37.7 & 43.2 & 81.6 & 46.8 & 42.9 & 73.3 & 90.3 & 21.4 & 56.7 & 84.4 & 55.0 & 62.9 & 82.5 & 63.2 \\
    Shen~\cite{Shen_2019_CVPR} &-&-&-&-&-&-&-&-&-&-&-&-&-&-&-&-&-&-&-&-& 63.5 \\
    $\text{Li}^{+}$ \cite{li2019weakly} &-&-&-&-&-&-&-&-&-&-&-&-&-&-&-&-&-&-&-&-& 67.9 \\
    $\text{C-MIL}^{+}$ \cite{Wan_2019} &-&-&-&-&-&-&-&-&-&-&-&-&-&-&-&-&-&-&-&-& 67.4 \\
    $\text{Yang }\textit{et al.}^{+}$ \cite{yang2019towards} & 82.4 & 83.7 & {72.4} & {57.9} & 52.9 & 86.5 & 78.2 & {78.6} & 40.1 & 86.4 & 37.9 & {67.9} & {87.6} & 90.5 & 25.6 & 53.9 & 85.0 & {71.9} & 66.2& 84.7 & 69.5 \\
    Arun~ \textit{et al.}\cite{Arun_2019} &-&-&-&-&-&-&-&-&-&-&-&-&-&-&-&-&-&-&-&-& 69.5 \\
    $\text{WSOD2}^{+}$ \cite{Zeng_2019_ICCV} &-&-&-&-&-&-&-&-&-&-&-&-&-&-&-&-&-&-&-&-& {71.9} \\
    $\text{MIST (Full)}^{+}$ \cite{ren-wetectron2020} & {91.7} & {85.6} & 71.7 & 56.6 & {55.6} & {88.6} & 77.3 & 63.4 & {53.6} & {90.0} & 51.6 & 62.6 & 79.3 & {94.2} & 32.7 & 58.8 & {90.5} & 57.7 & {70.9} & {85.7} & 70.9 \\
    $\text{IM-CFB}^{+}$ \cite{Yin_Deng_Zhou_Li_2021} &-&-&-&-&-&-&-&-&-&-&-&-&-&-&-&-&-&-&-&-& 69.6 \\
    GradingNet-C-MIL \cite{Jia_Wei_Ruan_Zhao_Zhao_2021} &-&-&-&-&-&-&-&-&-&-&-&-&-&-&-&-&-&-&-&-&  71.9 \\
     \textbf{OICR+REG (reproduce)} & 78.8&80.3&67.1&49.4&52.2&88.7&73.9&74.0&50.9&81.8&37.3&59.8&77.1&86.9&21.6&46.6&74.0&70.6&72.3&79.0&66.1\\
     \textbf{CASD (reproduce)} & 71.0&83.1&75.1&49.6&61.7&91.0&79.0&81.1&56.8&75.4&38.5&61.2&80.0&87.1&19.0&56.0&87.8&67.0&75.9&81.0&69.4\\
    \textbf{OICR+REG+W2N (Ours)} &87.4&86.0&69.7&50.8&59.8&89.8&88.4&86.9&37.5&86.5&26.0&69.8&84.0&95.1&31.6&57.6&78.12&75.6&85.8&77.3&\textbf{71.2(+5.1)}\\
    \textbf{CASD+W2N (Ours)}& 89.7&88.4&82.1&53.5&71.8&93.4&88.2&86.8&59.9&94.1&57.4&85.3&94.6&92.8&32.9&64.7&94.5&64.1&84.2&87.3&\textbf{78.3(+8.9)}\\
    \hline
    \textbf{WSOD with transfer learning}:\\
    $\text{LBBA}^{+}$ \cite{dong2021boosting}& 92.3 & 90.0 & 85.0 & 68.0 & 63.2 & 92.8 & 82.4 & 66.8 & 57.2 & 96.8 & 54.1 & 80.4 & 92.1 & 94.4 & 16.8 & 66.4 & 94.5 & 70.8 & 71.8 & 91.3 & \text{76.4} \\

     \textbf{LBBA (reproduce)} & 86.9&84.5&74.6&65.6&55.1&85.4&86.8&84.4&42.5&88.0&45.0&83.3&82.3&88,6&47.6&49.1&88.3&50.8&81.1&84.3&72.7\\

    \textbf{LBBA+W2N (Ours)}& 93.1&91.5&85.0&68.1&76.1&96.0&90.2&86.8&63.3&95.7&57.3&86.3&94.8&94.3&27.6&66.4&93.2&67.1&83.6&87.1&\textbf{80.2(+7.5)} \\
    \specialrule{.15em}{.05em}{.05em}
    \end{tabular}
    }}
    \label{table:per-cls-voc12-corloc}
    \end{table*}
\subsection{Comparison with State-of-the-arts}

First we compare our method with several state-of-the-art WSOD approaches in terms of mAP and CorLoc on PASCAL VOC 2007 and 2012~\cite{everingham2010pascal} reported by Table~\ref{table:per-cls-voc07},  ~\ref{table:per-cls-voc12}, ~\ref{table:per-cls-voc07-corloc} and ~\ref{table:per-cls-voc12-corloc}. Our all results are obtained with single-scale testing approach. Based on these results, we obtain the following observations: First, our W2N framework outperforms all WSOD baselines in terms of both mAP and CorLoc. Specifically, on PASCAL VOC 2007 dataset, it outperforms OICR+REG by  8.7\% mAP and 3.8\% CorLoc, outperforms CASD by 11.4\% mAP and 12.6\% CorLoc, and outperforms LBBA by 9.5\% mAP and  10.7\% CorLoc. Performance on PASCAL VOC 2012 also demonstrates favorable performance improvement. Second, our W2N outperforms all of the state-of-the-art WSOD methods as well as transfer learning based methods. Specifically, OICR+REG+W2N achieve 57.0\% mAP on PASCAL VOC 2007 test set, outperforming CASD \cite{huangCASD2020} by 0.2\% mAP which is the state-of-the-arts of pure WSOD method. CASD+W2N achieves 65.4\% mAP on PASCAL VOC 2007 test set, outperforming $\text{CaT}_{5}$ by 2.4\% mAP which is the state-of-the-arts of transfer learning based WSOD method. Moreover, LBBA+W2N obtains 68.6\% mAP  and 83.4\% CorLoc, which achieves a new state-of-the-arts for WSOD problem and catches up with the performance of fully supervised methods Faster R-CNN.

\begin{figure*}
   
    \begin{center}
    \includegraphics[width=1\linewidth]{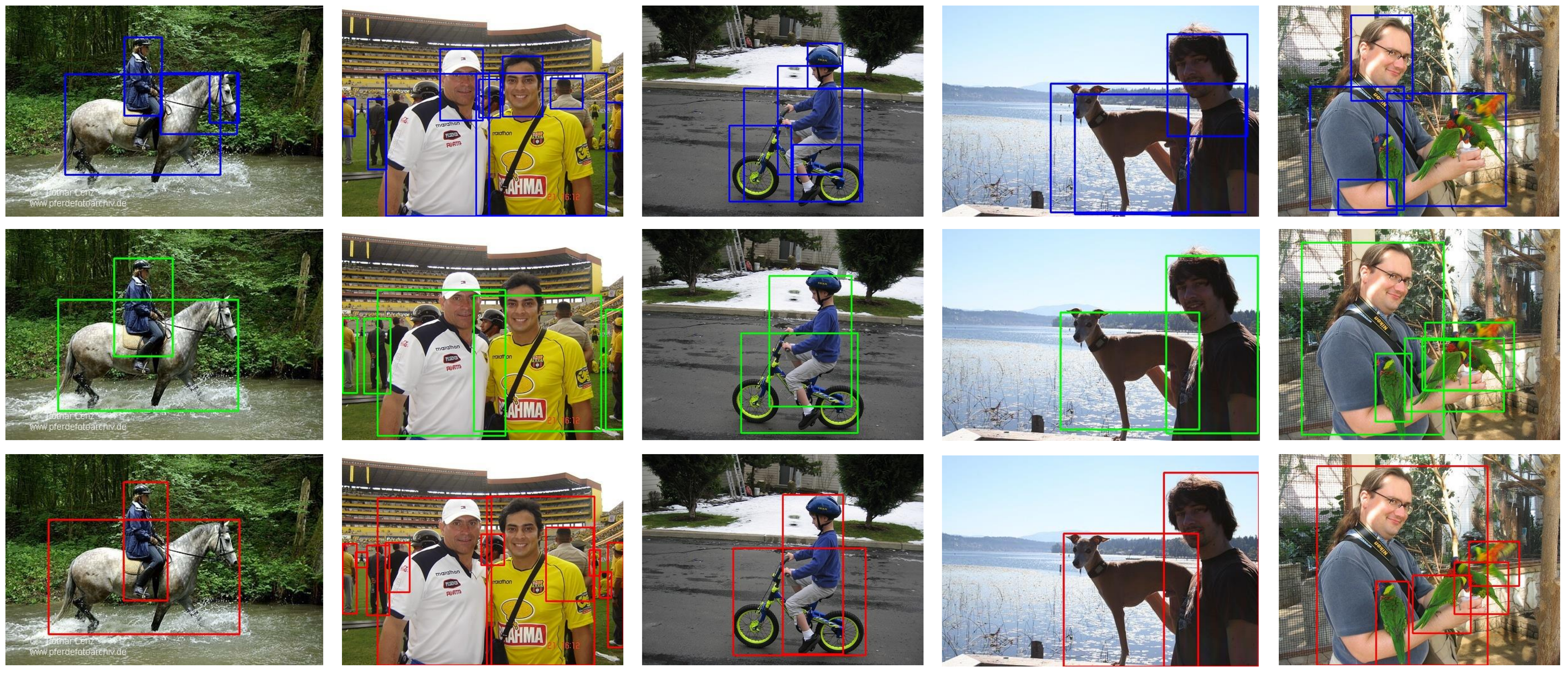}
    \end{center}

        \caption{Visualization results of our method on PASCAL VOC 2007 test set. \textbf{Top row}: detection results from LBBA. \textbf{Medium row}: detection results from LBBA with W2N. \textbf{Bottom row}: ground-truth annotations.}
  
    \label{fig:vis}
    \end{figure*}

Fig.~\ref{fig:vis} shows the visualization results of our method on PASCAL VOC 2007 test set. The top row is the detection results of LBBA (reproduce) , the medium row is the detection results of LBBA+W2N and the bottom row is ground-truth annotations. Obviously, with the help of W2N framework, the bounding box predicted by model have a better location performance . And the phenomenon that predict bounding boxes covering at the discriminative part have been eased. 

\subsection{Results on More Datasets}
We also deploy more experiments on MS-COCO 2017 dataset \cite{lin2014microsoft} and ILSVRC 2013  detection dataset \cite{deng2009imagenet} to prove the effectiveness of our method. MS-COCO dataset includes 80 categories with 118,287 training images and  5,000 validation images. ILSVRC 2013 detection dataset include 200 categories with 395,909 train images and 20,121 validation images. We choose \emph{OICR+REG} to implement the weakly supervised detector and conduct our proposed on these two datasets and the results are shown in Table~\ref{table:coco} and Table~\ref{table:imgn}. With the help of W2N training process, the OICR+REG outperform 10.8\% and 5.5\% in terms of AP50 and mAP on MS-COCO, 5.2\% AP50 on ILSVRC 2013, which demonstrates the generalization ability of our W2N framework.
\begin{table}[t!]
    \caption{Results of our method on MS-COCO 2017 validation  set.}
        \centering
            \footnotesize{
        \begin{tabular}{l |  c c c c c c}
        \specialrule{.15em}{.05em}{.05em}
        Methods  &  mAP  & AP50 & AP75 & $AP_S$ & $AP_M$ & $AP_L$ \\
        \hline
        OICR+REG \cite{tang2017multiple} (reproduce)& 9.8 & 20.8 & 7.9 & 1.4 & 9.2 & 17.7  \\
        
        OICR+REG+W2N & \textbf{15.3} & \textbf{30.0} & \textbf{13.9} & \textbf{4.9} & \textbf{18.5} & \textbf{24.6}  \\
        \hline
        CASD \cite{huangCASD2020} (reproduce)& 10.5 & 24.1 & 8.3 & 2.7 & 12.2 & 18.3  \\
        CASD+W2N & \textbf{15.9} & \textbf{33.3} & \textbf{13.4} & \textbf{5.6} & \textbf{18.4} & \textbf{27.2}\\
        \specialrule{.15em}{.05em}{.05em}
        \end{tabular}
        }
        \label{table:coco}
       
    \end{table}

\begin{table}[t!]
    \caption{Results of our method on ILSVRC 2013 detection validation set.}
        \centering
            \footnotesize{
        \begin{tabular}{l |  c }
        \specialrule{.15em}{.05em}{.05em}
        Methods  &  AP50 \\
        \hline
        OICR+REG \cite{tang2017multiple} & 17.4  \\
        OICR+REG+W2N & \textbf{22.6}  \\
        \hline
        CASD \cite{huangCASD2020} & 18.4  \\
        CASD+W2N & \textbf{27.9}  \\
        \specialrule{.15em}{.05em}{.05em}
        \end{tabular}
        }
        \label{table:imgn}
    
    \end{table}

\begin{table}[t]

\centering
\caption{Effect of two modules on VOC 2007.}

  \begin{tabular}{l | c | c | c | c | c }
  \specialrule{.15em}{.05em}{.05em}
  WSOD & FRCNN* & LA & SSL & ITER & mAP \\
  \hline
     \multirow{5}{*}{OICR+REG\cite{tang2017multiple}} &  & & & & 48.3 \\
                          & \checkmark & & & & 49.0 \\
                          &  & \checkmark & & & 49.9 \\
                          &  &  & \checkmark & & 56.1 \\
                          &  & \checkmark & \checkmark & & 56.8 \\
                        &  & \checkmark & \checkmark &\checkmark & \textbf{57.0} \\
  
    \specialrule{.15em}{.05em}{.05em}
    \multirow{5}{*}{CASD\cite{huangCASD2020}} &  & & & & 54.0 \\
                          & \checkmark & & & & 55.0 \\
                          &  & \checkmark & & & 55.6 \\
                          &  &  & \checkmark & & 62.0 \\
                          &  & \checkmark & \checkmark & & 62.7 \\
                        &  & \checkmark & \checkmark &\checkmark & \textbf{65.4} \\
  
    \specialrule{.15em}{.05em}{.05em}

    \multirow{5}{*}{LBBA\cite{dong2021boosting}} &  & & & & 59.1 \\
                          & \checkmark & & & & 59.4 \\
                          &  & \checkmark & & & 60.3 \\
                          &  &  & \checkmark & & 66.1 \\
                          &  & \checkmark & \checkmark & & 67.0 \\
                        &  & \checkmark & \checkmark &\checkmark & \textbf{68.6} \\
    
    \specialrule{.15em}{.05em}{.05em}
    \end{tabular}
  \label{table:com}

\end{table}

\begin{table}[t!]
  
    \caption{Training and Inference time comparison. }
    
    \centering
    \footnotesize{
    \begin{tabular}{l | c | c}
    \specialrule{.15em}{.05em}{.05em}
    Second phase methods & Training time(h/stage) & Inference time(s/img)\\
    \hline
    Faster R-CNN  & 0.5 & 0.06\\
    SoS & 7.2 & 0.06 \\
    W2N(Ours) & 7.8 & 0.06 \\
    \specialrule{.15em}{.05em}{.05em}
    \end{tabular}
    }
    \label{table:time}
   
    \end{table}

\begin{table}[t!]
   
    \caption{Effect of different backbone of W2N on VOC 2007. }

    \centering
    \footnotesize{
    \begin{tabular}{l | c }
    \specialrule{.15em}{.05em}{.05em}
    Methods & mAP \\
    \hline
    LBBA(VGG16) & 59.1  \\
    LBBA(VGG16)+W2N(VGG16) & 66.7 \\
    LBBA(VGG16)+W2N(Res50+FPN) & \textbf{68.6} \\
    \specialrule{.15em}{.05em}{.05em}
    \end{tabular}
    }
    \label{table:backbone}
  
    \end{table}
    
\subsection{More Ablation Study}

\noindent\textbf{Effect of two modules.} 
Table~\ref{table:com} shows the ablation study of each module on different WSOD
baselines.
Simply re-training Faster R-CNN(FRCNN*) with pseudo GT only outperforms \emph{OICR+REG} by 0.7\% mAP, \emph{CASD} by 1.0\% mAP and \emph{LBBA} by 0.7\% mAP  respectively. 
By introducing localization adaption and semi-supervised learning separately, these improvements respectively outperform the  \emph{OICR+REG} by 1.6\% and 7.8\% mAP; outperform the  \emph{CASD} by 1.6\% and 8.0\% mAP; and outperform the  \emph{LBBA} by 1.2\% and 7.0\% mAP; 
Furthermore, our full method combining these two modules can further improve the detection performance to 56.8\%,\quad 62.6\% and 67.0\% mAP respectively. \\

\noindent\textbf{Time cost of W2N:}  We show the training and inference time of W2N in Table \ref{table:time}.
Our framework increase slight training time in comparison to SoS \cite{sui2021salvage},  while having the same inference time as Faster R-CNN and SoS, which illustrates the high efficiency of our method.\\

\noindent\textbf{Effect of different backbones:}  For a fair comparison with \cite{zhong2020boosting,sui2021salvage}, we use VGG16 as the backbone of WSOD baselines and use Res50 with FPN as the backbone of target detector. 
Furthermore, we additionally conduct experiments by using Faster R-CNN with VGG-16 backbone as target detector to explore the performance that our method adapted on other backbones.
Table \ref{table:backbone} shows the effect of using different backbones as target detectors. W2N leads to better mAP on better backbone.\\

\begin{table*}[htb!]
  
    \caption{Comparison of different dataset split method on Pascal VOC 2007 for different WSOD baseline at iteration 0.}
    \centering
    \resizebox{\textwidth}{!}{
    \footnotesize{
    \begin{tabular}{l | c | c | c | c | c}
    \specialrule{.15em}{.05em}{.05em}
    Methods & {image-level split} & {instance-level split} & {two tasks instance-level split} & ideal split &mAP\\
    \hline
    OICR+REG+W2N & \checkmark &  &  & & 55.4 \\
    OICR+REG+W2N &  & \checkmark & & & 56.8 \\
    OICR+REG+W2N &  & & \checkmark & & \textbf{56.8} \\
    OICR+REG+W2N &  & &  & \checkmark & \textbf{61.4} \\
    \hline
    CASD+W2N & \checkmark &  & & &61.9\\
    CASD+W2N &  & \checkmark & & &62.6 \\
    CASD+W2N &  & & \checkmark & &  \textbf{62.7} \\
    CASD+W2N &  & &  & \checkmark &  \textbf{66.5} \\
    \hline
    LBBA+W2N & \checkmark &  &  & &65.4\\
    LBBA+W2N &  & \checkmark & & & 66.8 \\
    LBBA+W2N &  &  & \checkmark & &
    \textbf{67.0}\\
    LBBA+W2N &  &  &  & \checkmark &
    \textbf{72.8}\\
    \specialrule{.15em}{.05em}{.05em}
    \end{tabular}}}
    \label{table:splitt}
    \end{table*}
\section{Discussion}
\subsection{Ideal Data Split}
In this work we propose a \emph{hybrid-level dataset split} method, which aim to keep more pseudo label with high quality for subsequent SSOD training process. Theoretically, if we can keep more label with high quality, then the SSOD will perform better, which may be a important research direction for future work. To prove that the quality of labeled set will affect the performance of detector, we propose an simple \emph{ideal data split}  with the help of ground-truth dataset $\mathbb{X}_{gt}$. $\mathbb{X}_{gt} = \{\mathbf{I}, \{\mathbf{S}_{gt}\}\}$, $\mathbf{S}_{gt} = (\mathbf{b}_{gt},c_{gt})$, where $\mathbf{S}_{gt}$ denotes a ground-truth box annotation with box coordinate $\mathbf{b}_{gt}$ and category $c_{gt}$. Specifically, given the $i$ th pseudo bounding box label $\mathbf{S}^{(i)} = (\mathbf{b}^{(i)},c^{(i)})$ of a training image $\mathbf{I}$ with location $\mathbf{b}^{(i)}$ and category $c^{(i)}$ predicted by weakly-supervised detector, calculate the Intersection over Union (IoU) between the $\mathbf{S}^{(i)}$ and every ground-truth of $\{\mathbf{S}_{gt}\}$ whose category $c_{gt}$ as same as $c$. Then we keep the $i$ th pseudo box label in labled set if there is at least one ground-truth box label has IoU with $\mathbf{S}^{(i)}$ higher than 0.5. And other pseudo label will be discard. Note that introducing the instance-level ground-truth annotation is only for illustrating the effect of clean label and it is not allowed in WSOD task.

We also deploy experiments for ideal split method for every WSOD baseline and Table~\ref{table:splitt} shows the result of ideal data split method. Obviously, comparing with other data split method which we proposed, ideal split further improves performance a lot. For example, for LBBA+W2N, ideal split outperforms 5.8\% mAP than two tasks instance-level split. Hence we believe that it is worth to explore how to design a more effective data split method in future work.



\begin{figure}[htbp]
\begin{minipage}[t]{0.47\linewidth}
\centering
    \includegraphics[scale=0.4]{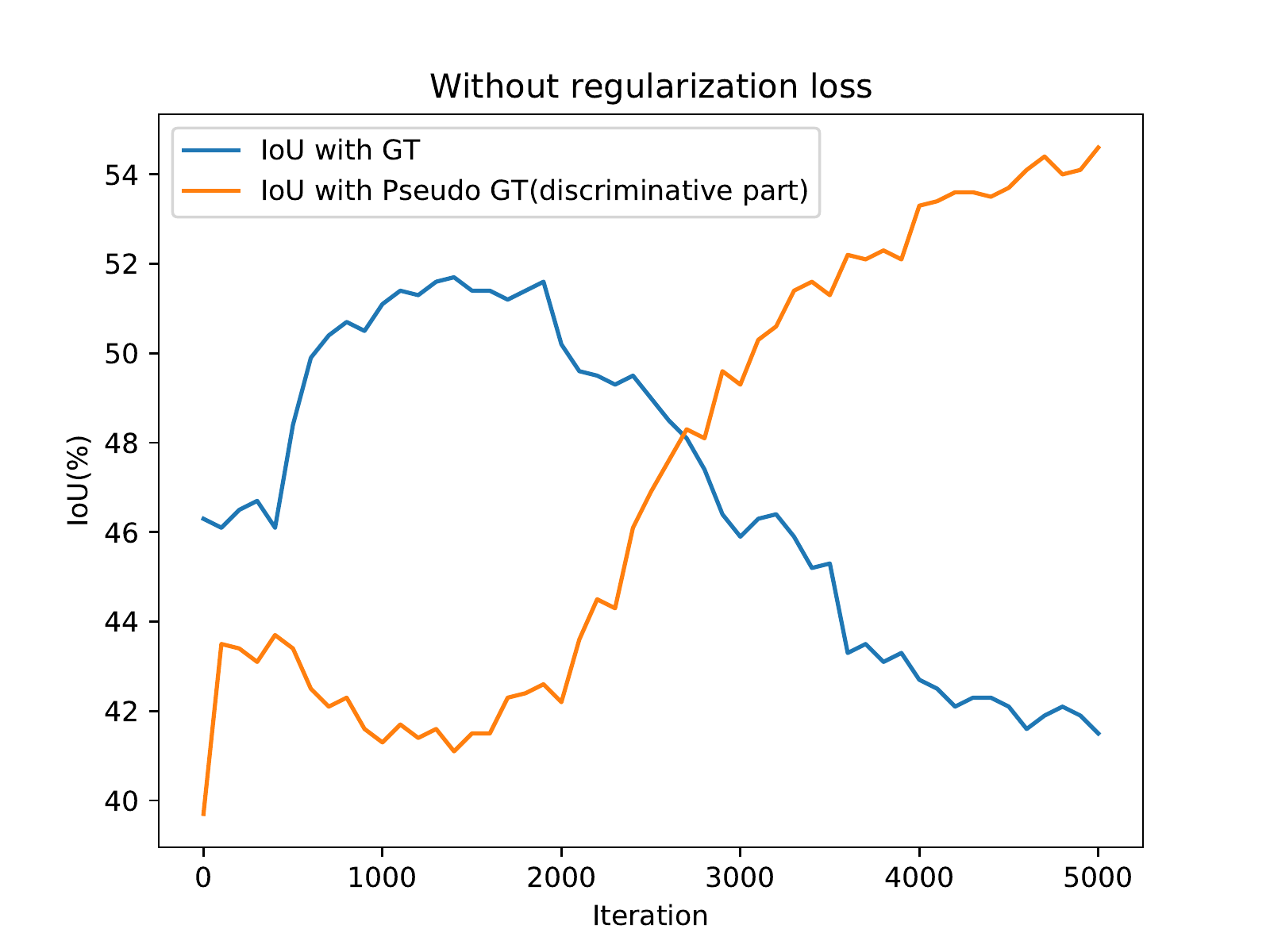}


\end{minipage}
\hfill
\begin{minipage}[t]{0.47\linewidth}
\centering
    \includegraphics[scale=0.4]{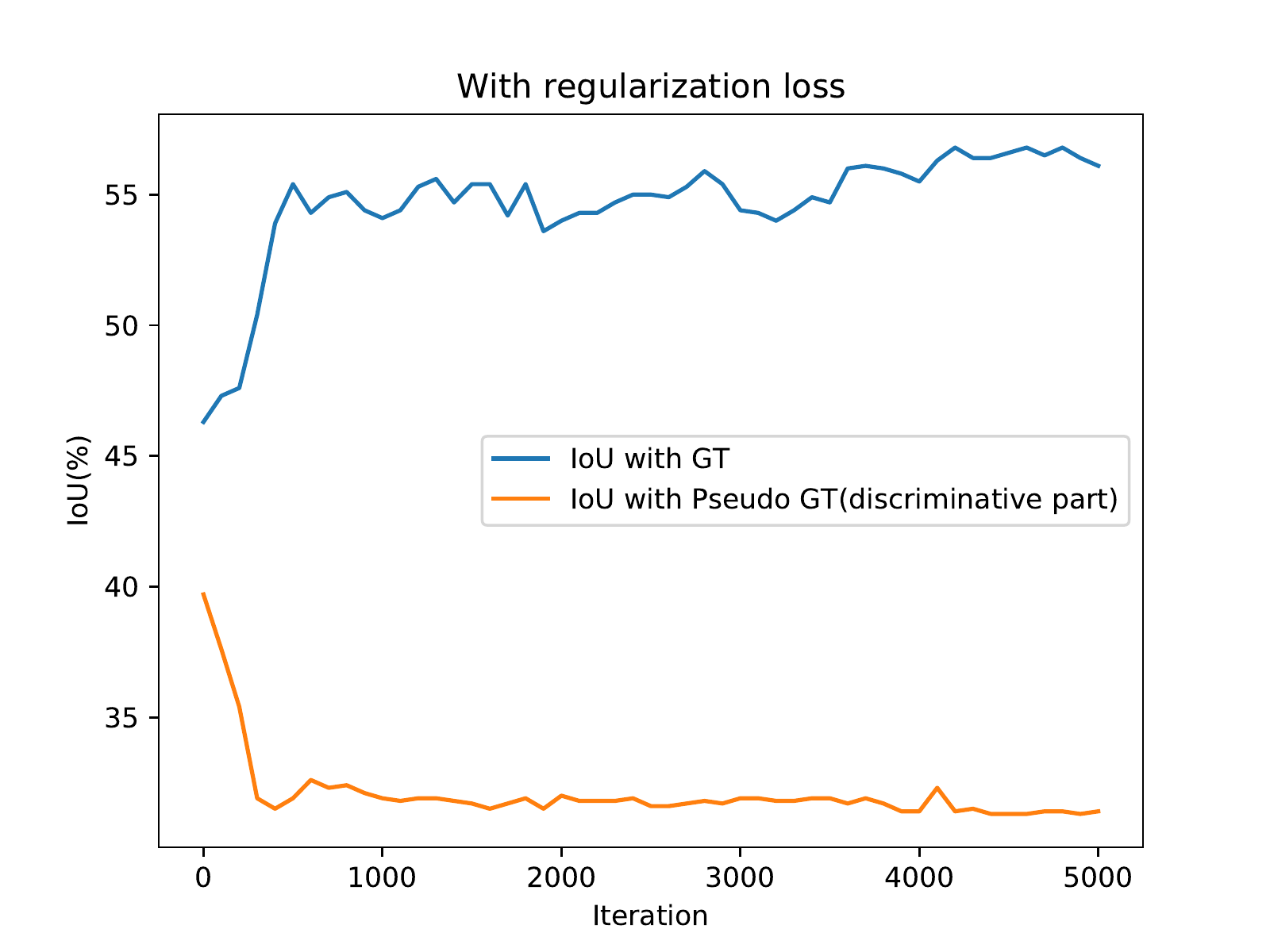}


\end{minipage}
\caption{Curves of IoU trends between estimated box and GT/pseudo GT.}
\label{fig:qa}   
\end{figure}

\subsection{Quantitative analyze of ``outer proposals'' observation:}
\vspace{-0.5em}
We propose the localization adaptation module and the regularization loss based on the early stage learning characteristic of bounding box regression. To prove this observation and investigate the effectiveness of regularization loss, we conduct an quantitative experiment by plotting IoU curves between proposals and GT/pseudo GT.
In each iteration, we randomly sample $N$ (\eg, 30) outer proposals around each pseudo GT with ``discriminative part problem", obtain the current bbox regression w/ and w/o regularization loss, and calculate the IoU metric between estimated box and GT/pseudo GT (Fig~\ref{fig:qa}). 
Without regularization loss, the outer proposals first regress to the GT at early learning stage, but finally regress to pseudo GT (the left sub-figure).
In comparison, with regularization loss, the outer proposals can be regressed consistently towards GT along with the training iterations(the right sub-figure). Thus, the above result can provide an empirical support to our observation as well as the effectiveness of regularization loss.







\vspace{-1em}
\subsection{Discussion of Sui \etal}
\vspace{-0.5em}
Sui \etal\cite{sui2021salvage}. proposed a novel WSOD framework named SoS, which is first to adopt SSOD in WSOD task. There are several differences between this work and ours.  

\textbf{First}, in overall, SoS only adopt SSOD method to enhance the performance of WSOD,  while our work formulates the multi-phase weakly supervised object detection problem as a noisy-label object detection problem. Learning with noisy labels has been widely studied for image classification, where the noise is image-level and is on classification labels. 
In contrast, noisy label remains less investigated for object detection due to that: {\color{blue}(i)} noise of pseudo label on localization is also inevitable, and {\color{blue}(ii)} both classification and localization label noise are instance-level instead of image level. 
Here we present \emph{Location Adaption module} and \emph{Hybrid-Level Dataset Split} for handling these two issues, which are novel for incorporating noisy label learning with WSOD. In addition, we believe that as the noisy-label learning develops, we can absorb more idea from it and design more better performing model for WSOD task.

\textbf{Second}, SoS applies \cite{li2019dividemix} for object detection task directly, while we think more about the characteristic of (weakly-supervised) object detection task in terms of the noise distribution. For example, In \emph{Location Adaption Module}, we focus on the ``discriminative part'' issue of WSOD and analyze the change of regression of bound boxes during training, which inspires us to propose the regularization loss to make the location performance better. In \emph{Semi-supervised Learning Module}, we propose that both classification and localization label noise are instance-level instead of image level. So, we calculate classification and localization loss respectively for every instance, which can screen more high-quality samples for training. 

\textbf{Third}, our W2N outperforms 2.7\% mAP and 1.2\% mAP than SoS with the same WSOD baseline(CASD) at single scale testing on PASCAL VOC 2007 and 2012 dataset respectively. Note that we don't adopt any training tricks \eg multi input training strategy and PGF proposed in \cite{sui2021salvage}.

\begin{figure*}

    \begin{center}
    \includegraphics[width=1.\linewidth]{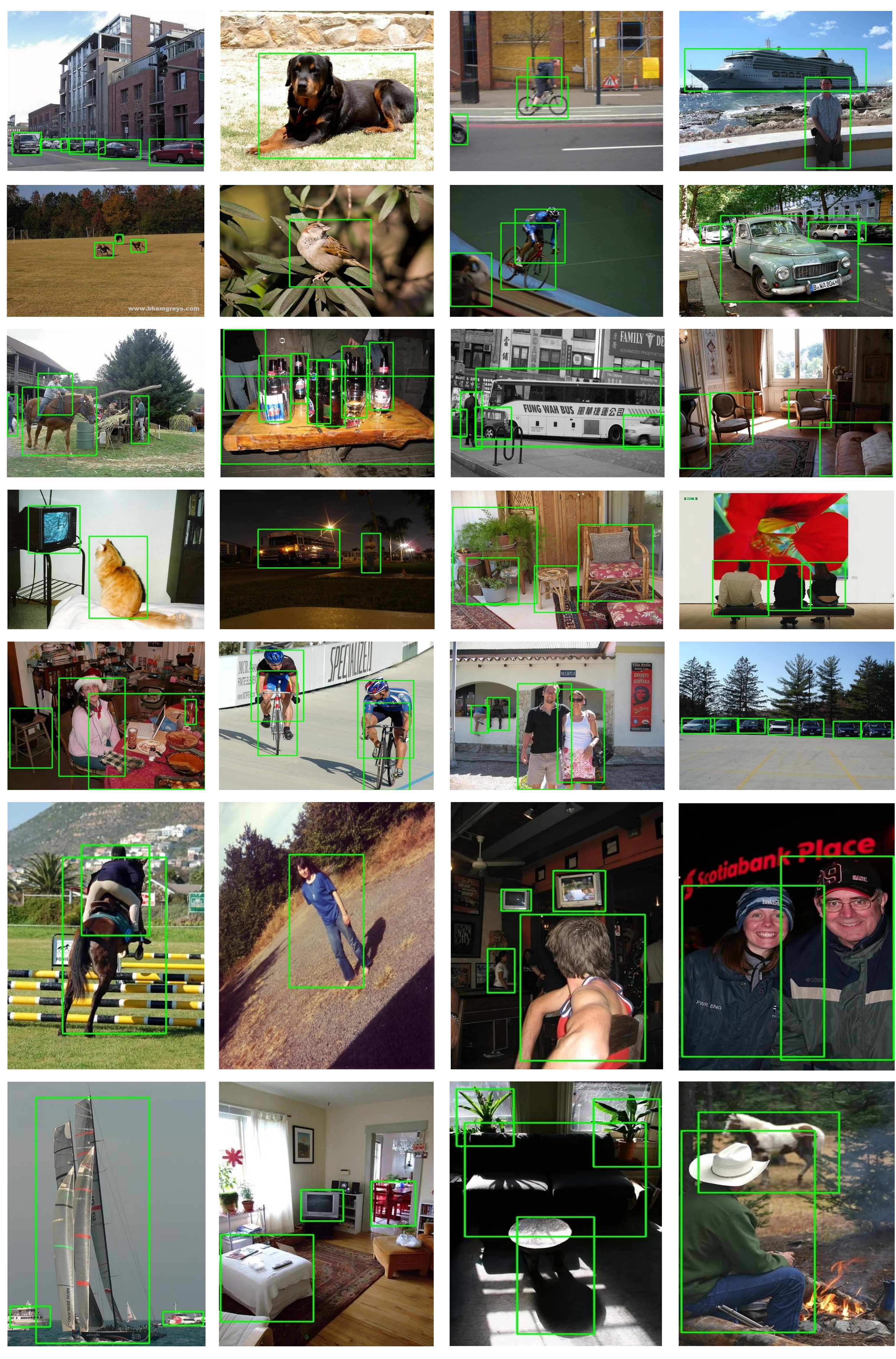}
    \end{center}

        \caption{More visualization results of LBBA+W2N on PASCAL VOC 2007 test set.}
  
    \label{fig:vis_all}
    \end{figure*}  
    
\clearpage

\end{document}